\documentclass[preprint,11pt]{elsarticle}

\makeatletter
\def\input@path{{elsevier/}}
\makeatother

\usepackage[T1]{fontenc}
\usepackage{mathptmx}
\usepackage[margin=2.5cm]{geometry}
\usepackage{setspace}
\usepackage[font=footnotesize]{caption}
\usepackage{graphicx}
\usepackage{amsmath,amssymb}
\usepackage{booktabs}
\usepackage{makecell}
\usepackage{tabularx}
\usepackage{xcolor}
\usepackage[hidelinks]{hyperref}
\usepackage{placeins}
\usepackage{adjustbox}

\graphicspath{{figures/}{../figures/}}

\journal{arXiv}
\makeatletter\def\ps@pprintTitle{\let\@oddhead\@empty\let\@evenhead\@empty\let\@oddfoot\@empty\let\@evenfoot\@empty}\makeatother



\newsavebox\tabbox
\newenvironment{fitwidth}{\begin{lrbox}{\tabbox}}{\end{lrbox}\typeout{TABWIDTH table \thetable\space natural \the\wd\tabbox\space line \the\linewidth}\ifdim\wd\tabbox>\linewidth\resizebox{\linewidth}{!}{\usebox\tabbox}\else\usebox\tabbox\fi}
\usepackage{etoolbox}
\patchcmd{\pprintMaketitle}{\elsaddress\par\vskip36pt}{\elsaddress\par\vskip12pt}{}{\PackageWarning{main}{title patch failed}}
\patchcmd{\MaketitleBox}{\elsaddress\par\vskip36pt}{\elsaddress\par\vskip12pt}{}{\PackageWarning{main}{title patch failed}}
\patchcmd{\pprintMaketitle}{\unvbox\keybox\par\vskip10pt\fi\hrule\vskip12pt}{\unvbox\keybox\par\vskip10pt\fi\vskip12pt}{}{\PackageWarning{main}{keyword rule patch failed}}
\makeatletter\g@addto@macro\normalsize{\setlength\abovedisplayskip{5pt plus 2pt minus 2pt}\setlength\belowdisplayskip{5pt plus 2pt minus 2pt}\setlength\abovedisplayshortskip{2pt plus 2pt}\setlength\belowdisplayshortskip{5pt plus 2pt minus 2pt}}\makeatother
\makeatletter
\renewcommand\section{\@startsection {section}{1}{\z@}{12\p@ \@plus 4\p@ \@minus 3\p@}{6\p@ \@plus 4\p@ \@minus 3\p@}{\normalsize\bfseries\boldmath}}
\renewcommand\subsection{\@startsection{subsection}{2}{\z@}{8\p@ \@plus 4\p@ \@minus 3\p@}{3\p@ \@plus 4\p@ \@minus 3\p@}{\normalfont\normalsize\itshape}}
\makeatother
\begin{document}

\begin{frontmatter}

\title{CAPTOR: text-encoder-free grounding by counting caption co-occurrence over the visual words of a frozen encoder}

\author[tau]{Elad Yoshai}
\author[tau]{Natan T. Shaked\corref{cor}}
\cortext[cor]{Corresponding author.}
\ead{nshaked@tau.ac.il}
\address[tau]{School of Biomedical Engineering, Faculty of Engineering,
Tel Aviv University, Tel Aviv 6997801, Israel}

\begin{abstract}
Grounding a text query in an image needs a matrix scoring every term
against every patch. Vision-language models compute it with an aligned text encoder, which many image encoders, such as histopathology-related models trained on slides alone, lack. The Caption-Anchored Patch-Term Occurrence Representation
(CAPTOR) estimates the matrix without a text encoder. It clusters a frozen image encoder's patch tokens into visual words and describes each captioned image by its share of
patches in each word. A term signature is the standardized difference of those shares between images whose captions mention the term and the rest. The matrix
factorizes into a term-by-word table and a query-independent image factor, so an
image is encoded once and a new term adds one table row. Against every training-free baseline, CAPTOR leads on Common Objects in Context (COCO), Pascal Visual Object Classes (VOC) and Cityscapes. On COCO, scored on the cells one object owns,
it reaches $0.565$ mean intersection over union (mIoU) against $0.502$. On all three, it is also level with a text alignment trained on the same backbone, which
needs gradients, a text encoder and several times the captions. On held-out breast-cancer slides, it exceeds the strongest baseline scoring patches with a pathology text encoder by $0.160$ mIoU. Read through CAPTOR, UNI, an encoder
trained on slides alone, exceeds such baselines by $0.132$ on a common grid. Counting can therefore give a text interface to any image encoder trained without language, wherever captioned images exist.

\end{abstract}


\end{frontmatter}

\section{Introduction}
\label{sec:intro}

A vision and language model connects a text query to an image through a
term-by-patch compatibility matrix, which holds one number for each term and each
patch saying how strongly that term applies there. Segmentation maps, retrieval
scores and the weights of text-image cross-attention are all readings of it. The
matrix is normally produced by a text encoder that turns each term into a query
vector, which is then compared with a key at every patch.

That route needs a text encoder aligned to the image encoder, and many image
encoders have none. Aligning one needs large sets of paired images and text and
the training to use them, which natural images supply and specialized domains
often do not. Histopathology has the gap. Some pathology foundation models were
trained on image and caption pairs and can be queried with text~\cite{conch}.
Others were trained on slide patches alone~\cite{uni,virchow,gigapath} and carry
no language, so no text query can be formed against them. A trained alignment
such as Talk2DINO~\cite{talk2dino} does not remove this need, since it maps a
pretrained text encoder onto each image encoder and must be trained again for
every encoder it serves.

In this paper, we propose a new technique that builds the matrix without a text encoder. We cluster a frozen image encoder's patch tokens into $K$ visual words and describe each captioned image by
the fraction of its patches in each word. For every term we then measure how
those fractions differ between images whose captions mention the term and images
whose captions do not. We express that difference in units of what shuffled
captions produce for the same term. We call the resulting table of terms against
words, and the method that builds it, the Caption-Anchored Patch-Term Occurrence
Representation (CAPTOR). The compatibility matrix is that table composed with the
patch-to-word assignment.

Counting how caption words co-occur with quantized image regions goes back to automatic image annotation~\cite{mori1999,duygulu,barnard}, which named images and their regions with words before vision-language models and foundation encoders existed. CAPTOR uses counting for a different purpose, to build for a frozen foundation encoder the term-by-patch matrix that a vision-language model forms with a text encoder. It differs from that work in what it estimates, how it normalizes the estimate and what the estimate is used for. The first difference is the estimate, counted signatures that a permutation null
puts on one scale across terms of very different frequency, so that the best term
at a patch can be read as its label. Frequency correction, more than the choice of statistic, is what matters, and the permutation null performs it best among the corrections tried. The second is the interface. The matrix
factorizes into a term-by-word table times an image factor that does
not depend on the query, which is exactly the term-patch scoring half of a
cross-attention layer. An image is therefore encoded once. Querying a term reads one row of a small table, and adding a term the captions support takes one counting pass rather than a pass through a text encoder. The third is
where it applies. Because no text encoder enters at any stage, the construction
gives a text interface to image encoders that have none.

We compare CAPTOR with the training-free methods built on Contrastive
Language-Image Pre-training (CLIP)~\cite{clip}, which, like CAPTOR, take no
gradient. Each is run from its released code and, except on VOC, also given the same propagation as a control.
CAPTOR leads them on Common Objects in Context (COCO)~\cite{coco}, scored on the
grid cells that one object owns, and on Pascal Visual Object Classes
(VOC)~\cite{voc}. Under the standard full-image protocol, it also leads them on
Cityscapes~\cite{cityscapes} and is level with them on COCO-Object. It is also level on COCO, VOC and Cityscapes with Talk2DINO, which
trains an alignment on top of CLIP's text encoder with several times the captions,
while taking no gradient, consulting no text encoder and encoding each image once. In their released settings rather than the matched ones, Talk2DINO scores higher on COCO and VOC, and ProxyCLIP and Talk2DINO score higher on COCO-Object. Both build on the $400$ million image--caption pairs of CLIP, where CAPTOR counts at most $30{,}000$ COCO captions. In histopathology, on patients held out from every choice, CAPTOR on the text-trained
encoder CONCH exceeds every baseline that reads CONCH's text encoder without
borrowing CAPTOR's visual words, and UNI, trained on slides alone, read through
CAPTOR exceeds those baselines on breast and is level with them on lung.

The contributions of this paper are the following.
\begin{itemize}\itemsep2pt
\item A text-encoder-free and gradient-free estimator of the term-by-patch
 compatibility matrix, counted from captions and normalized by a permutation null,
 with joint words and joint estimation that resolve where visual words and terms
 overlap (Section~\ref{sec:method}).

\item The factorization $A = S\Psi$, which replaces the term-patch scoring of
 cross-attention for the terms a caption corpus supports. An image is encoded once,
 a term is one row, and the hard, soft and propagated scoring rules are three
 choices of one image factor (Sections~\ref{sec:method} and~\ref{sec:attn}).

\item An evaluation against baselines reproduced to their published numbers and, except on VOC, given the same propagation, on a matched protocol and on the standard full-image
 protocols, with paired intervals throughout (Sections~\ref{sec:nat}
 and~\ref{sec:standard}).

\item Text-queryable patch description for pathology encoders that have no text
 encoder, which on breast-cancer tissue exceeds dense scoring with a trained pathology text encoder and on lung is level with it, with a specialist vocabulary acquired from a few thousand
 captions (Section~\ref{sec:path}).
\end{itemize}

\section{Related work}
\label{sec:related}

Previous open-vocabulary grounding methods form their query with a text encoder (Table~\ref{tab:families}). CLIP
and its pathology descendants~\cite{clip,plip,conch} align whole images with
captions, and dense predictions are recovered from them without training by
reading their patch tokens against the text embedding~\cite{maskclip,sclip}.
ProxyCLIP~\cite{proxyclip} replaces CLIP's own attention with the patch affinity
of a self-supervised encoder and so reads two backbones. A trained route learns
the missing alignment instead. Talk2DINO~\cite{talk2dino} maps CLIP's text
embeddings onto DINOv2~\cite{dinov2} patch features, dino.txt~\cite{dinotxt}
fits a text encoder against a frozen DINOv2, and Segment Anything Model~3
(SAM~3)~\cite{sam3} is trained on millions of concept labels, and
Section~\ref{sec:s-sam3} of the supplement compares it with CAPTOR. All of them encode the term with a text encoder when a query arrives.

\begin{table}[!tbp]\centering
\caption{Comparison of previous families of methods. Every method, CAPTOR included, uses development labels only to select its settings.}
\label{tab:families}
\begin{fitwidth}
\begin{tabularx}{\linewidth}{>{\raggedright\arraybackslash}Xcccc}
\toprule
Method family & \makecell{Text\\encoder} & Gradients & \makecell{Segmentation\\labels in training} & \makecell{Cost of a\\new term} \\
\midrule
Dense CLIP scoring~\cite{maskclip,sclip} & Yes & No & No & Encoder pass \\
Several backbones~\cite{proxyclip} & Yes & No & No & Encoder pass \\
Trained alignment~\cite{talk2dino,dinotxt,zoomshot} & Yes & Yes & No & Encoder pass \\
Mask supervised~\cite{sam3} & Yes & Yes & Yes & Encoder pass \\
\midrule
CAPTOR & \textbf{No} & \textbf{No} & \textbf{No} & \textbf{Caption counting} \\
\bottomrule
\end{tabularx}
\end{fitwidth}
\end{table}

An image encoder trained
without language can be made to answer text queries by training a mapping from its
features into a space already aligned with text. Zoom-shot~\cite{zoomshot}, for
example, maps a vision encoder's features into CLIP's latent space. In pathology, foundation models trained on slides alone~\cite{uni,virchow,gigapath,phikon} are becoming standard tools~\cite{mahmood_survey}. The models that can be queried with text are trained on paired images and captions~\cite{conch,plip}, and their fine-grained alignment is still being improved by distillation~\cite{vdpath}. Each of these routes needs paired data, gradients and
a text encoder. In contrast, CAPTOR needs captioned images and no gradient-based training, and it leaves the
image encoder untouched.

Binding caption words to
quantized image regions is where automatic image annotation began. Mori et al.~\cite{mori1999} divide images into a grid, quantize the
regions, transfer each caption's words to every region and read each word's
distribution off the counts. Later work replaced counting with estimation,
through words-and-blobs translation models fitted by
expectation-maximization~\cite{duygulu,barnard} and cross-media relevance
models~\cite{jeon2003}. In each of them, a word is linked to clusters of
image regions according to how often the two appear together in captioned images. CAPTOR shares
their caption supervision but differs from them in four ways. Its statistic is a
standardized contrast between images that name a term and images that do not,
divided by a permutation null so that terms of any frequency share one scale. Its
visual words are clusters of the patch tokens of a modern foundation model, where the
earlier methods clustered hand-crafted color and texture features. Its table factorizes the
term-patch scoring of cross-attention and supports dense segmentation with no text
encoder. Its reweighting lets the terms of a caption compete for patches, as
expectation-maximization does, but applies once to its own table.
Section~\ref{sec:nat} below compares the two statistics on the same visual words.

\section{Method}
\label{sec:method}

\begin{figure*}[!tbp]\centering
\includegraphics[width=\linewidth]{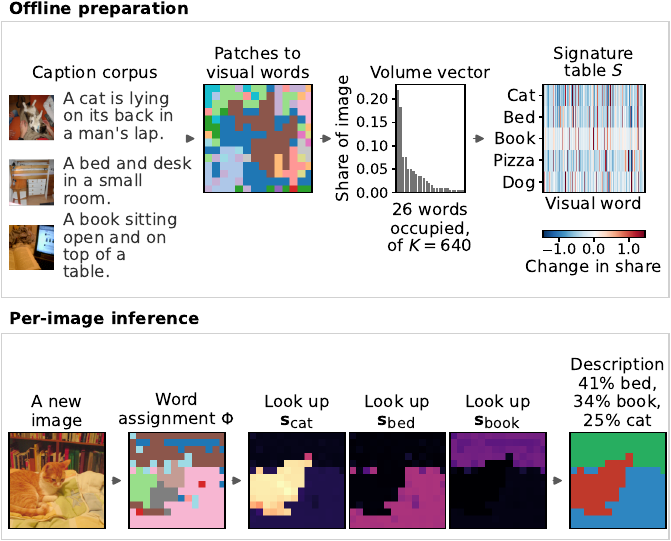}
\caption{The method, drawn from its own artifacts rather than as a schematic.
\textbf{Offline preparation.} Patch features of $15{,}000$ captioned COCO images
are clustered into $K=640$ visual words, and each term signature is counted from
them. The table is the factor $S$ of Equation~\eqref{eq:attn}, one row per term
and one column per word. Red means a word covers more of an image when captions
mention the term and blue less, in multiples of what shuffled captions produce.
\textbf{Per-image inference.} One forward pass assigns each patch of a new image
to its nearest word on the encoder's $14\times14$ grid, by the rule of
Equation~\eqref{eq:volume} that described the corpus. That word map is $\Phi$, and
scoring a term at a patch reads the row of the term in $S$ at the word of the patch, the
product $S\Phi$ entry by entry. The table holds five terms, the row shows the three
the image contains, and the three maps share one color scale.}
\label{fig:pipeline}
\end{figure*}

CAPTOR has two stages. An offline stage builds a table from a caption corpus, and
an online stage answers queries about a new image from that table
(Figure~\ref{fig:pipeline}). No gradient is taken and no text encoder is consulted
at any point.

\subsection{The compatibility matrix as two factors}
\label{sec:m-matrix}

Let an image be divided into $P$ patches and let a vocabulary hold $T$ terms. The
\emph{compatibility matrix} $A \in \mathbb{R}^{T \times P}$ has one entry per
term and patch, and $A_{tp}$ says how strongly term $t$ applies at patch $p$. A
vision and language model forms it entry by entry as
$A_{tp} = \langle g(t), h(x_p)\rangle$, where $g$ is a text encoder that maps the
term to a query vector, $h$ projects the image encoder token $x_p$ of patch $p$
to a key, and both vectors have dimension $d$. We build the same matrix as a
product of two factors,
\begin{equation}
A \;=\; S\,\Phi, \qquad
S \in \mathbb{R}^{T\times K}, \quad \Phi \in \{0,1\}^{K\times P}.
\label{eq:attn}
\end{equation}
The \emph{term factor} $S$ is a table of the $T$ terms against $K$ visual words,
and the \emph{image factor} $\Phi$ records which word each patch belongs to.
Section~\ref{sec:m-words} builds both from a frozen image encoder and a caption
corpus, Section~\ref{sec:m-props} generalizes $\Phi$ to any image factor that does
not depend on the query, and Section~\ref{sec:m-refine} sharpens the estimate of
$S$ where visual words and terms overlap.

\subsection{Counted signatures on a common scale}
\label{sec:m-words}

Let $f$ be a frozen vision transformer (ViT)~\cite{vit} that turns an image into
$P$ patch tokens $x_1, \dots, x_P \in \mathbb{R}^{d}$, each normalized to unit
length. The table is built from a \emph{caption corpus}, $N$ images that each
carry a caption and no other annotation. The $k$-means algorithm~\cite{kmeanspp}
runs over the patch tokens of all $N$ images together, and its $K$ centroids
$c_1, \dots, c_K \in \mathbb{R}^{d}$ are the \emph{visual words}. They are fitted
once and held fixed, so a word means the same thing in every image. Each patch is
assigned to its nearest word, and image $i$ is summarized by the fraction of its
patches in each word,
\begin{equation}
\begin{aligned}
w(p) &\,=\, \arg\min_{k \in \{1,\dots,K\}}\|x_p - c_k\|,\\[2pt]
v^{(i)}_k &\,=\, \tfrac{1}{P}\bigl|\{\,p : w(p) = k\,\}\bigr|,
\qquad \textstyle\sum_{k} v^{(i)}_k = 1.
\end{aligned}
\label{eq:volume}
\end{equation}
We call $v^{(i)} \in \mathbb{R}^{K}$ the \emph{volume vector} of image $i$. It
records how much of the image each word occupies and not where. The vocabulary
size $K$ sets how finely the method can distinguish what it names.

The table is filled by counting. Let $M_t$ be the set of corpus images whose caption contains
term $t$, matched as Section~\ref{sec:setup} states. The \emph{signature} of $t$
is the change in the volume of each word between the images in $M_t$ and the rest, in
units of the standard deviation of that word across the corpus,
\begin{equation}
s_t[k] \,=\,
\frac{\operatorname{mean}_{i \in M_t} v^{(i)}_k
 - \operatorname{mean}_{i \notin M_t} v^{(i)}_k}
 {\delta_k},
\qquad
\delta_k \,=\, \operatorname{sd}_{i=1}^{N}\, v^{(i)}_k .
\label{eq:signature}
\end{equation}
So $s_t[k]$ measures how much of an image word $k$ tends to occupy when term $t$
is mentioned, compared with when it is not. The caption enters only through
membership of $M_t$, so no text encoder is involved, and the quantity measured is a
property of images that a text encoder cannot observe from language alone.

\begin{figure*}[!tbp]\centering
\includegraphics[width=\linewidth]{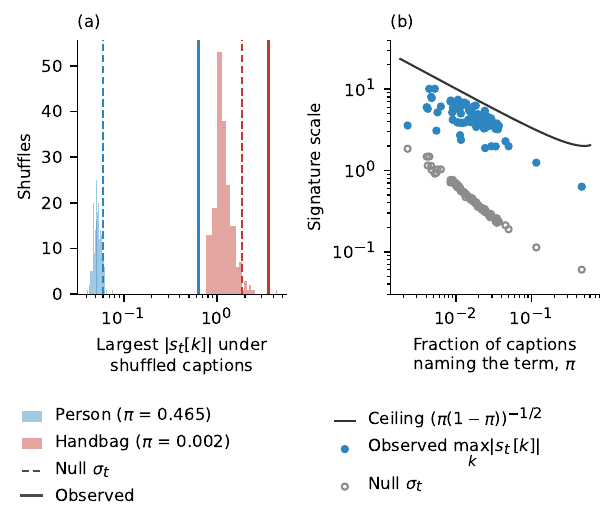}
\caption{The permutation null and the ceiling on COCO, drawn with one vocabulary
and the hard assignment. \textbf{(a)} The permutation null of
Equation~\eqref{eq:null} for a common term and a rare one, where $\pi$ is the
fraction of captions that name a term. ``Person'', with $\pi = 0.465$, reaches
$0.631$ against a null of $0.060$. ``Handbag'', named in $34$ of $15{,}000$
captions, with $\pi = 0.002$, reaches $3.57$ against $1.85$. On the raw scale the
rare term is more than five times larger. Measured in multiples of its own null,
however, it is the weaker of the two, at $1.9$ times its null against $10.5$ for
``person''. \textbf{(b)} The largest raw
signature of every term, its null and the ceiling of Equation~\eqref{eq:ceiling}. Both rise as a
term becomes rare, and no term exceeds $0.84$ of its ceiling.}
\label{fig:mechanism}
\end{figure*}

The largest coordinate $\max_k |s_t[k]|$ is large even for a term that carries no
signal, because among $K$ noisy coordinates some are large by chance alone. We
therefore measure how large it becomes when the captions carry no information
about the images. To do so, we shuffle the captions. Each shuffle, or draw, keeps every image volume vector and the number $|M_t|$ of images that mention the
term, but picks those images at random. We repeat the draw $B$ times. With
$s^{(b)}_t$ the signature computed on draw $b$ and $m_b$ its largest coordinate,
the \emph{permutation null} $\sigma_t$ is the 95th percentile,
$\mathrm{P}_{95}$, of the $B$ maxima,
\begin{equation}
\begin{aligned}
\sigma_t &\,=\, \mathrm{P}_{95}\bigl\{\, m_1, \dots, m_B \,\bigr\},
\qquad B = 200,\\
m_b &\,=\, \max_k \bigl|s^{(b)}_t[k]\bigr| .
\end{aligned}
\label{eq:null}
\end{equation}
A term whose own $\max_k |s_t[k]|$ exceeds $\sigma_t$ is bound more strongly than
chance reaches on nineteen draws out of twenty, and we say it \emph{clears its
null} (Figure~\ref{fig:mechanism}a).

Even above their nulls, signatures of different terms cannot be compared
directly. With $\pi_t$ the fraction of captions that contain $t$, the
point-biserial correlation between mentioning $t$ and the volume of word $k$ is
$r_t[k] = s_t[k]\sqrt{\pi_t(1-\pi_t)}$, and since a correlation lies in $[-1,1]$,
\begin{equation}
|s_t[k]| \;\le\; \big(\pi_t(1-\pi_t)\big)^{-1/2} .
\label{eq:ceiling}
\end{equation}
The ceiling depends on the term frequency and on nothing else. It equals $2$ at
$\pi_t=\tfrac12$ and grows without bound for rare terms, so the largest raw score
at a patch would be decided largely by how often each term is mentioned. Every
signature is therefore divided by its own null,
\begin{equation}
\tilde{s}_t[k] \,=\, s_t[k] \,/\, \sigma_t ,
\label{eq:nullscale}
\end{equation}
which expresses each term in multiples of what shuffled captions produce for that
same term, a scale computed from the corpus alone. Dividing term scores by a positive number cannot change their order, so the division
leaves the ranking of patches for each term exactly as it was. On COCO that ranking
reaches a mean area under the receiver operating characteristic curve (AUC) of
$0.983$ with or without the division. The whole effect of the division is therefore on
the comparison between terms, where it raises the mean intersection over union
(mIoU) from $0.429$ to $0.511$ (Table~\ref{tab:coco-abl}). Much of this gain is frequency correction alone. On all COCO images at $K=1280$, Welch's $t$, which divides by the analytic standard error, raises the raw signature from $0.452$ to $0.526$ mIoU, and the null reaches $0.541$ (Table~\ref{tab:estimators}). From here on, $s_t$
denotes $\tilde{s}_t$ on natural images. Pathology is different. Its five tissue
terms are all frequent, each named in $4$ to $17\%$ of the captions, so their
nulls lie within a factor of about two of one another. On COCO, the nulls of
``person'' and ``handbag'' alone differ by a factor of thirty. The division
therefore has little to correct in pathology. On the $41$ development patients, the division even lowers mIoU slightly
at every vocabulary size, from $0.452$ to $0.435$ at $K=80$. The selection rule of
Section~\ref{sec:setup} therefore leaves it off in pathology. On the held-out
patients the choice barely matters, at $0.497$ without the division and $0.494$
with it.
Figure~\ref{fig:mechanism}a and~b show the null and the ceiling on COCO.

With every signature on this common scale, the two factors of
Equation~\eqref{eq:attn} are complete. The image factor records the word of each
patch and the term factor holds the signatures,
\begin{equation}
\begin{aligned}
\Phi_{kp} &\,=\,
\begin{cases}
1 & \text{if } w(p) = k,\\
0 & \text{otherwise,}
\end{cases}\\[4pt]
S_{tk} &\,=\, \tilde{s}_t[k].
\end{aligned}
\label{eq:factors}
\end{equation}
so $A_{tp} = S_{t,w(p)}$ and scoring a patch against a term is one table lookup.

\subsection{Reading the matrix}
\label{sec:m-score}

The same table answers how strongly a term applies to a whole image and how
strongly it applies at each patch.

To score a whole image, we compare its composition with the term signature.
The share $v_k$ of each word in the image is standardized with statistics of the
caption corpus and weighted by the signature,
\begin{equation}
\text{score}(v, t) \,=\, \sum_{k=1}^{K} s_t[k]\, z_k ,
\qquad
z_k \,=\, \frac{v_k - \mu_k}{\delta_k},
\qquad
\mu_k \,=\, \operatorname{mean}_{i=1}^{N}\, v^{(i)}_k ,
\label{eq:score}
\end{equation}
where $v$ is the image volume vector, and $\mu_k$ and $\delta_k$ are the mean and
standard deviation of the share of word $k$ over the corpus. Both are stored with the
table, so no statistic of an evaluation image enters the score. The whole
signature is used because an object or a tissue usually spans several words.
Since $v_k$ is the fraction of the image patches assigned to word $k$,
$v_k = \frac{1}{P}\sum_{p}\Phi_{kp}$, the score can be written with the two
factors of Equation~\eqref{eq:attn},
\begin{equation}
\text{score}(v,t) \,=\, \sum_{k=1}^{K} \frac{S_{tk}}{\delta_k}
\Bigl(\frac{1}{P}\sum_{p=1}^{P}\Phi_{kp} - \mu_k\Bigr).
\label{eq:scorefact}
\end{equation}
It therefore reads the same row of $S$ and the same assignment $\Phi$ as the
patch scores.

At a single patch the reading is simpler. A patch belongs to exactly one word $w(p)$, so
\begin{equation}
\text{score}(p, t) \,=\, s_t[w(p)] \,=\, A_{tp}.
\label{eq:patchscore}
\end{equation}
Because each row of $S$ is divided by its own null, the scores of different terms
at one patch share a scale, and the highest-scoring term is the patch label.

\subsection{The image factor}
\label{sec:m-props}

Under the hard assignment every patch is read through its single word. Two
patches in the same word therefore receive identical scores for every term, and a
patch near the boundary between two words can be assigned to the wrong one. The
construction does not require the hard assignment. It requires only an image
factor that is computed from the image alone, without the query, and has one row
per word. Writing such a factor as $\Psi \in \mathbb{R}^{K \times P}$, the
compatibility matrix becomes
\begin{equation}
A \;=\; S\,\Psi,
\label{eq:general}
\end{equation}
so Equation~\eqref{eq:attn} is the case $\Psi = \Phi$, and the score of term $t$ at
patch $p$ is $A_{tp} = \sum_{k} S_{tk}\Psi_{kp}$.
$\Psi$ is computed once per image before any term is known and is reused for
every term. A query therefore reads a single row of $S$. A term estimated on its
own, as in Section~\ref{sec:m-words}, adds one row to $S$ without changing the
others, and the joint estimate of Section~\ref{sec:m-refine} is the one exception.
We use two assignments, and a propagation that can follow either of them.

The first assignment is the hard assignment $\Phi$ of
Equation~\eqref{eq:factors}. The second, the \emph{soft assignment}, lets a patch
share its weight among its $m$ nearest words. With $N_m(p)$ the set of the $m$
words whose centroids lie nearest to the token $x_p$,
\begin{equation}
\Psi_{kp} \,=\, \alpha_k(p) \,=\, \frac{e^{\,\beta\, x_p^{\!\top} c_k}}
{\sum_{k' \in N_m(p)} e^{\,\beta\, x_p^{\!\top} c_{k'}}}
\ \ \text{if } k \in N_m(p),
\qquad
\Psi_{kp} \,=\, 0 \ \ \text{otherwise},
\label{eq:softscore}
\end{equation}
where $x_p^{\!\top} c_k$ is the similarity between the patch token and the centroid
of word $k$, and the sharpness $\beta$ sets how strongly the weight concentrates on
the nearest word. The score of a term at the patch is then the weighted average of
the signatures of those $m$ words. Setting $m=1$ returns the hard assignment.
Development data select $m=16$ and $\beta=10$ on natural images, and $m=1$ in
pathology (Table~\ref{tab:hyper} in the supplement).

The \emph{propagation} lets the patches of one object support one another. A
patch can be misread on its own, but other patches of the same object usually look
like it. Each patch's scores are therefore replaced by a weighted average over the
patches that resemble it, wherever they lie in the image. With $x_p^{\!\top} x_q$
the cosine similarity of two unit-length tokens and $\bar{a}$ its average over all
pairs of patches in the image,
\begin{equation}
W_{pq} \,=\, \frac{u_{pq}}{\sum_{q'} u_{pq'}},
\qquad
u_{pq} \,=\,
\begin{cases}
\exp\!\big(\kappa\, x_p^{\!\top} x_q\big)
  & \text{if } x_p^{\!\top} x_q > \theta\,\bar{a},\\
0 & \text{otherwise,}
\end{cases}
\label{eq:affinity}
\end{equation}
so patch $q$ counts only when its similarity to $p$ exceeds $\theta$ times the
image average, and the gain $\kappa$ sets how much more a very similar patch
counts. We use $\theta = 1.2$ and $\kappa = 3$ in both domains, fixed once and
never tuned. The propagated scores are $A W^{\top} = S\,\Psi W^{\top}$, so
propagation only replaces the image factor $\Psi$ by $\Psi W^{\top}$, still
computed from the image alone, and leaves $S$ unchanged. ProxyCLIP~\cite{proxyclip}, the
strongest training-free baseline, propagates in the same way but takes $W$ from a
second model, whereas CAPTOR takes both the assignment and $W$ from the one
encoder it reads. Propagation uses only the image, not the term, so we apply it to the
scores of every compared method, and the comparison measures only how terms are
scored.

\subsection{Estimating the table where words and terms overlap}
\label{sec:m-refine}

Counting over visual words has two weaknesses. One visual word may cover patches
that show different things, and a caption credits every term it names with the
whole image, including the other things in it. Joint words and trust weights
address the first, and caption-conditioned reweighting or joint estimation the
second. None of them uses a text encoder, a gradient or an annotation, each
reduces to plain counting at one setting, and each is used only where development
data favor it.

\emph{Joint words} give each patch a finer label. The $k$-means algorithm started
from different random seeds draws the boundaries between words in different
places, so each vocabulary wrongly merges different pairs of patches. We fit $J$ such vocabularies to the same tokens and label each patch by the
combination of its $J$ words,
\begin{equation}
\ell(p) \,=\, \big(w_1(p), \dots, w_J(p)\big),
\qquad
\text{score}(p,t) \,=\, s_t[\ell(p)],
\label{eq:joint}
\end{equation}
where $w_j(p)$ is the word of patch $p$ in the $j$th vocabulary and $s_t[\ell]$ is
Equation~\eqref{eq:signature} counted over these combined labels. For example, if
three vocabularies put a cat's ear and a dog's ear in the same word and the fourth
separates them, their joint words differ. Two patches thus share a joint word only
when every vocabulary merges them. Averaging the $J$ score maps instead would let
the three vocabularies outvote the fourth. Like words, joint words are computed
once per image.

The combinations are many, so a joint word is often seen in only a few captioned
images, too few for a reliable signature. Its score is then blended with the
average of the $J$ single-vocabulary scores,
\begin{equation}
\text{score}(p,t) \,=\, \gamma_{\ell}\, s_t[\ell(p)]
 \;+\; (1-\gamma_{\ell})\,\frac{1}{J}\sum_{j=1}^{J} s^{(j)}_t\!\left[w_j(p)\right],
\qquad
\gamma_\ell \,=\, \frac{n_\ell}{n_\ell+\lambda},
\label{eq:trust}
\end{equation}
where $\ell = \ell(p)$, $s^{(j)}_t$ is the signature from the $j$th vocabulary
alone and $n_\ell$ is the number of captioned images that contain $\ell$. The
\emph{trust weight} $\gamma_\ell$ moves from the average toward the joint word as
$n_\ell$ grows, and the two count equally when $n_\ell = \lambda$. With
$\lambda = 3$, chosen for natural images, a joint word seen once gets weight
$0.25$ and one seen thirty times $0.91$
(Figure~\ref{fig:mechsupp}, Section~\ref{sec:s-method} of the supplement). Pathology chooses a very large $\lambda$,
which leaves only the average.

For the second weakness, take the caption ``a dog is holding a frisbee''.
Equation~\eqref{eq:signature} credits ``frisbee'' with every patch of the
image, the dog and the grass included. \emph{Caption-conditioned reweighting}
instead divides each patch among the terms the caption names, according to how
well each term fits it. With $T(i)$ the terms named by the caption of image $i$ and
$\text{score}(p,t)$ the score of a first pass of the table, patch $p$ counts
toward $t \in T(i)$ with weight
\begin{equation}
r_{ip}(t) \,=\, \varepsilon + (1-\varepsilon)\,
\frac{e^{\,\tau\,\text{score}(p,t)}}
     {\sum_{t' \in T(i)} e^{\,\tau\,\text{score}(p,t')}},
\label{eq:em}
\end{equation}
where $\tau$ sets how sharply the terms compete and the floor $\varepsilon$ keeps a
term from being cut off by an early mistake. A patch of the dog thus counts mostly
toward ``dog'' and little toward ``frisbee''. The signature of $t$ is then
recounted from the weighted share of each word in each image $i \in M_t$,
\begin{equation}
v^{(i,t)}_k \,=\, \frac{\sum_{p} r_{ip}(t)\,[\,w(p) = k\,]}{\sum_{p} r_{ip}(t)},
\label{eq:emvolume}
\end{equation}
where $w(p)$ is the word of $p$ in the vocabulary being counted and $[\cdot]$ is
one when its condition holds.
Images that do not name $t$, the deviations $\delta_k$ and the nulls $\sigma_t$
stay unweighted. No annotation is used, and a caption that names a single term
leaves its image unchanged.

\emph{Joint estimation} solves the same problem for all terms at once. It treats
the word shares of each captioned image as a sum of one contribution per named term,
fitted to all images together. Terms named together thus share the explanation,
and each keeps only what the others cannot explain.
Suppose a visual word whose patches show table surfaces covers $30\%$ of every image whose caption names
``dining table'' and $2\%$ of every other image. Bottles usually stand on tables,
so suppose also that three of every four captions naming ``bottle'' name ``dining
table'' too. Separate counting therefore credits ``bottle'' with the word. The prediction instead
finds that naming ``dining table'' adds $0.28$ to the share and naming ``bottle''
adds nothing, which fits every image. In matrix form, $Y \in \{0,1\}^{N \times T}$ has
$Y_{it} = 1$ when the caption of image $i$ names term $t$, and the rows of
$V \in \mathbb{R}^{N \times K}$ are the volume vectors. A bar
subtracts each column mean. The prediction is $\bar{V} \approx \bar{Y} B$, where
$B_{tk}$ is how much naming term $t$ adds to the share of word $k$ given the other
named terms. $B$ minimizes the squared error
$\|\bar{V} - \bar{Y} B\|^2$ plus a ridge penalty $\lambda_J N \|B\|^2$, which keeps
it stable when terms often co-occur, giving
\begin{equation}
\hat{s}_t[k] \,=\, \frac{B_{tk}}{\delta_k},
\qquad
B \,=\, \bigl(\bar{Y}^{\top}\bar{Y} + \lambda_J N\, I\bigr)^{-1}\,\bar{Y}^{\top}\bar{V},
\label{eq:jointest}
\end{equation}
with $I$ the $T \times T$ identity. Each $\hat{s}_t$ is then divided by
its null, computed as in Equation~\eqref{eq:null} but shuffling whole caption
sets so that terms named together stay together. $Y$ may also list unqueried terms,
such as the COCO classes outside the twenty VOC classes, to take words a queried term would
otherwise absorb. A dataset uses at most one of the
two corrections, the reweighting on COCO and the joint estimate on VOC.

\section{Experimental setup}
\label{sec:setup}

On natural images, the CAPTOR table is built from the captions of a
$15{,}000$-image subset of the COCO training split~\cite{coco}, which contains no
evaluation image. CAPTOR is evaluated on the $5{,}000$ images of the COCO validation
split and on the Pascal Visual Object Classes (VOC) validation set~\cite{voc}. Two
standard benchmarks are added. COCO-Object adds a
background class to the eighty COCO object classes, as in~\cite{sclip}, and
Cityscapes~\cite{cityscapes} contributes its $500$ validation images. In pathology
the table is built from the $32{,}463$ PathCap~\cite{pathcap} figures whose
caption mentions one of $28$ tissue terms. Three of these terms are merged
into others, leaving $25$. Evaluation uses the pixel annotation of the Breast
Cancer Semantic Segmentation (BCSS) dataset~\cite{bcss} and of
LUAD-HistoSeg~\cite{luadhistoseg}. Section~\ref{sec:s-setup} of the supplement lists every dataset,
split and setting (Tables~\ref{tab:datasets} and~\ref{tab:hyper}) and shows that
no evaluation image nearly copies a PathCap figure (Table~\ref{tab:overlap}).

Natural images are scored under two protocols. The \emph{matched protocol} scores
COCO and VOC on a $14\times14$ grid over the center crop of each image, the grid
both backbones produce without up-sampling, so every method sees the same
pixels. A cell is scored only when one object class covers more than half of it.
This keeps $162{,}215$ of the $490{,}000$ cells of the $2{,}500$ COCO reporting
images and $82{,}058$ of the $284{,}004$ cells of all $1{,}449$ VOC validation
images. Leaving out background isolates how each method scores terms, since each method decides background by its own rule. The \emph{standard protocol} scores every pixel of the
whole image, background included, with each baseline at its released settings and
checked against its published number (Section~\ref{sec:standard}).

Natural images use DINOv2 ViT-B/14 with
registers~\cite{dinov2,registers} at an input size of $196$ pixels, and CLIP
ViT-B/16~\cite{clip}. In pathology, CONCH~\cite{conch} is trained on
image-caption pairs, while UNI~\cite{uni} and Phikon~\cite{phikon} are trained on
slide patches alone and have no text encoder. All encoders stay frozen.

The natural-image baselines run the authors' released code of
SCLIP~\cite{sclip}, ProxyCLIP~\cite{proxyclip} and Talk2DINO~\cite{talk2dino},
together with dense CLIP scoring, which compares each patch token with the text
embedding of a term. Each uses its published prompts and scores a class at every patch by the best of its fixed names, which reads no annotation. The reported scores of CAPTOR include the propagation of
Equation~\eqref{eq:affinity}, so every baseline except on VOC is also scored with the same propagation, computed from the same DINOv2 tokens, and reported margins are against these propagated baselines. On VOC the baselines are compared as released. In pathology, the baselines score each patch
by comparing the CONCH text embedding of a term with the patch features, following
MaskCLIP~\cite{maskclip}, SCLIP, and the smoothing of CLIP-DINOiser~\cite{clipdinoiser}
along patch similarities. Each is given the same propagation, and each method,
CAPTOR included, may use a per-term standardization of its scores that reads no
annotation, chosen on development patients.

CAPTOR takes no gradient and uses no text encoder, and captions are its only
supervision for the table. The CLIP-based baselines read a text encoder trained on
about $400$ million image-caption pairs. Talk2DINO trains its alignment on the
$82{,}783$ captioned images of COCO 2014, more than five times the images CAPTOR
counts. Every evaluation set is split into a development half and a
reporting half, with each image, or in pathology each patient, in exactly one of
them. For BCSS, $41$ of the $103$ patients form the development half and the
other $62$ are scored once. The development half and its segmentation labels
choose the seven settings of Table~\ref{tab:hyper} and the background threshold
of the standard protocols. They also choose which components of
Section~\ref{sec:m-refine} each dataset uses, and the eight terms that need their
full name because the bare word is too general, such as ``stop sign''. Every headline number is measured
once on the reporting half, and curves and ablations that use all
images are marked as such. As a check, $K$ is also chosen from captions alone, by
how well signatures from one half of the corpus rank the images of the other half.
This rule selects $K=2560$ on natural images and $K=320$ in pathology, against $640$
and $80$ chosen with the development labels. With these values, CAPTOR reaches $0.569$ mIoU on the COCO reporting half, against $0.565$, and $0.441$ on the held-out BCSS patients, against $0.497$. A caption names a term when it
contains one of the names of the term in WordNet~\cite{wordnet}, a lexical database of
English, such as ``plane'' or ``jet'' for ``airplane''. Longer terms are matched
first, so ``teddy bear'' is not also read as ``bear''.

Every difference between two methods is computed from their paired predictions. Its $95\%$ interval comes from a percentile
bootstrap~\cite{efron} that resamples the largest independent unit of each
dataset, images on COCO, VOC and Cityscapes and patients on BCSS and
LUAD-HistoSeg. A difference whose interval excludes zero is called resolved, and
two methods whose difference is not resolved are called level, which means only that the data do not resolve a difference, not that the methods are equivalent. The $p$-values are not corrected for multiple
comparisons, so each should be read on its own.

\section{Results}
\label{sec:results}

\subsection{Open-vocabulary segmentation on COCO}
\label{sec:nat}
\label{sec:general}

We evaluate open-vocabulary segmentation on the COCO validation split under the
matched protocol, against the released code of each baseline. Every method labels
each scored cell with one of the same COCO object classes. Of the $80$ classes,
$78$ are named in at least $25$ of the $15{,}000$ corpus captions, and all $78$
clear their permutation null. The other two, ``toaster'' and ``hair drier'', are
named in $15$ and $0$ captions, so they are left out for every method.
Section~\ref{sec:s-coco} of the supplement adds qualitative results and
ablations (Figure~\ref{fig:coco}, Tables~\ref{tab:coco-k} and~\ref{tab:coco-steps}).

CAPTOR leads every training-free baseline on both measures, whether each is taken
as released or given the same propagation CAPTOR uses (Table~\ref{tab:cocoseg}).
It reaches $0.565$ mIoU and $0.768$ accuracy against $0.481$ and $0.670$ for
ProxyCLIP as released and $0.502$ and $0.690$ with the propagation, while reading
one backbone where ProxyCLIP reads two. The paired difference against the
propagated ProxyCLIP is $+0.064$, with a $95\%$ interval from $+0.044$ to
$+0.081$. Talk2DINO reads the same backbone but trains its term side, and appears at its
released configuration and at our input size.

\begin{table*}[!tbp]\centering
\caption{Open-vocabulary segmentation on COCO under the matched protocol, on
the $2{,}500$ reporting images and $78$ counted classes. \emph{Caption pairs} is how
much image--caption data each text side was learned from: the $15$k captions CAPTOR
counts, the $400$M pairs of CLIP, and for Talk2DINO also $83$k COCO captioned
images. The propagated columns give each baseline the propagation of
Equation~\eqref{eq:affinity}, and differences, with paired intervals that resample
images, are taken against them. The last two rows are references, not matched:
Talk2DINO as released reads $448$ instead of $196$ pixels and adds a pixel-adaptive
refinement, and the probe is fitted to the development labels.}
\label{tab:cocoseg}
\setlength{\tabcolsep}{2.8pt}
\begin{fitwidth}
\begin{tabularx}{\linewidth}{>{\raggedright\arraybackslash}Xcccccc}
\toprule
Method & \makecell{Caption\\pairs} & mIoU & \makecell{Accu-\\racy} & \makecell{mIoU,\\propa-\\gated} & \makecell{Accuracy,\\propa-\\gated} & \makecell{CAPTOR\\$-$ this,\\propa-\\gated} \\
\midrule
CAPTOR & $15$k & $0.565$ & $\mathbf{0.768}$ & $0.565$ & $\mathbf{0.768}$ & \\
Talk2DINO~\cite{talk2dino} & $400$M$\,{+}\,$$83$k & $\mathbf{0.566}$ & $0.746$ & $\mathbf{0.566}$ & $0.757$ & \makecell{$-0.001$\\ $[-0.017, +0.011]$} \\
ProxyCLIP~\cite{proxyclip} & $400$M & $0.481$ & $0.670$ & $0.502$ & $0.690$ & \makecell{$+0.064$\\ $[+0.044, +0.081]$} \\
SCLIP~\cite{sclip} & $400$M & $0.409$ & $0.593$ & $0.468$ & $0.650$ & \makecell{$+0.098$\\ $[+0.080, +0.115]$} \\
Dense CLIP scoring & $400$M & $0.142$ & $0.280$ & $0.190$ & $0.394$ & \makecell{$+0.375$\\ $[+0.355, +0.394]$} \\
\midrule
Talk2DINO, as released & $400$M$\,{+}\,$$83$k & $0.588$ & $0.764$ & & & \\
Supervised probe & Labels & $0.587$ & $0.866$ & & & \\
\bottomrule
\end{tabularx}
\end{fitwidth}
\end{table*}

Against Talk2DINO, accuracy and mIoU give different results. On accuracy, the share of
annotated patches given the right term, CAPTOR reaches $0.768$ against $0.757$ for
Talk2DINO given the same propagation and $0.746$ without it, ahead by $+0.011$ with
an interval from $+0.002$ to $+0.020$ that excludes zero. On mIoU, which averages over terms
so that a rare term counts as much as a common one, the difference is $-0.001$
with an interval containing zero, so the difference is not resolved. The two methods also make different kinds of errors. Averaged over the $78$
classes, $0.683$ of the patches CAPTOR assigns to a term are correct, against
$0.646$ for Talk2DINO, a paired difference of $+0.037$ with an interval from
$+0.019$ to $+0.048$. Talk2DINO, in turn, recovers more of the patches that belong to a term, $0.810$ against $0.740$ for CAPTOR. CAPTOR is therefore the more
precise method and Talk2DINO the more complete one.  As released, at $448$ pixels with a pixel-adaptive refinement, Talk2DINO leads by $0.023$ mIoU at similar accuracy, about $0.014$ from the refinement, which does not help CAPTOR, and $0.009$ from the larger input (Table~\ref{tab:t2d} in the supplement). A supervised linear probe on the same frozen tokens, fitted with the development
labels to show what these features allow, reaches $0.587$ mIoU, and CAPTOR attains
$96\%$ of it without training on labels.

The estimator itself accounts for part of this result. With the same visual words
and the same scoring, the difference of means of Equation~\eqref{eq:signature}
divided by its null reaches $0.541$ mIoU at $K=1280$ over all $5{,}000$ images, and
without the null the signature falls to $0.452$. The null also improves five of the seven other statistics we tried, all except Welch's $t$, which is already scaled, and a ridge fit (Table~\ref{tab:estimators} in the supplement).
One of them is the co-occurrence share of Mori et al.~\cite{mori1999}, the
fraction of the patches of a word whose caption names the term. In the reported
configuration, with the settings of each estimator chosen on the development half, it
reaches $0.555$ mIoU against $0.567$ for the difference of means of CAPTOR, a difference
of $-0.012$ with an interval from $-0.019$ to $-0.003$. The co-occurrence share
raises accuracy from $0.771$ to $0.806$ by favoring frequently captioned terms, at
the cost of rare classes (Table~\ref{tab:morireported} in the supplement).

CAPTOR loses most to Talk2DINO on objects that captions usually mention together, labeling $55\%$ of microwave cells as oven, $54\%$ of potted plants as vase and $60\%$ of cups as dining table. At $K=1280$ the joint estimate of Section~\ref{sec:m-refine} raises correct microwave cells from $0.25$ to $0.34$ but lowers development mIoU from $0.550$ to $0.548$. The development half therefore keeps the separate estimate on COCO and selects the joint one on Pascal VOC, where this confusion dominates (Table~\ref{tab:joint} in the supplement).

CAPTOR also transfers to Pascal VOC without new captions. The vocabulary, the volume vectors
and every signature are counted on COCO captions, and VOC supplies only images and
annotation. With the configuration chosen on the development half
(Table~\ref{tab:voc}), CAPTOR reaches $0.829$ mIoU on the reporting half. On VOC, most errors of the separate estimate
are bottles, screens and potted plants labeled ``dining table'', and the joint
estimate reduces them, a gain of $+0.022$ with an interval from $+0.009$ to
$+0.034$. CAPTOR leads ProxyCLIP at $0.801$ by $+0.027$, with an interval from
$+0.005$ to $+0.051$, and SCLIP at $0.750$ by $+0.079$. It is level with Talk2DINO
at our input size, at $-0.011$ with an interval from $-0.030$ to $+0.008$, while
Talk2DINO as released leads it by $0.022$. Section~\ref{sec:s-voc} of the supplement gives every configuration
(Tables~\ref{tab:voc} and~\ref{tab:voct2d}) and a qualitative comparison
(Figure~\ref{fig:voc}).

\providecommand{\pending}[1]{\textbf{[#1]}}

\subsection{The standard protocols}
\label{sec:standard}

The matched protocol isolates how each method scores terms, but its numbers
cannot be compared with the numbers the baselines publish. Under it, a baseline
that scores low might simply have been run wrongly, and nothing would reveal it. We therefore also score every method on the
standard full-image protocols, where each baseline must first reproduce its
published number. Every image is read whole, background is a class of its own,
and every baseline keeps its released resize, window, prompts and background
rule. CAPTOR reads each image at its own resolution with the propagation applied
over the whole image. Like SCLIP, it calls a pixel background when one of SCLIP's
scene names, such as ``sky'', ``wall'' and ``road'', wins it, each name
counted from the captions as one more term, or when no object term reaches $2.5$
times its permutation null, the one setting chosen on the development half.

On COCO-Object CAPTOR reaches $0.350$, above SCLIP and level with ProxyCLIP, both given the same propagation, at $0.346$ and $0.353$, a difference of $-0.003$ with an
interval from $-0.015$ to $+0.014$. ProxyCLIP as released reaches $0.373$ and
Talk2DINO $0.448$. CAPTOR names more object pixels correctly than any baseline,
but labels background far from any object as the object usually seen there, such
as ``skis'' on snow or ``surfboard'' on water (Section~\ref{sec:s-cobj}, Table~\ref{tab:cobj} and Figure~\ref{fig:cobjqual} in the supplement).

\begin{figure*}[!tbp]\centering
\includegraphics[width=\linewidth]{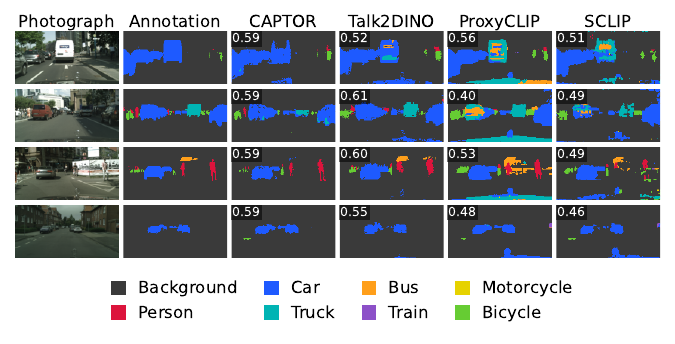}
\caption{Cityscapes at its full resolution of $2048\times1024$, on four typical
images, one per row. Among the reporting-half images whose
annotation contains at least two of the seven object classes, we take the four whose
CAPTOR score lies closest to the median, $0.59$. They show what a reader should expect rather than our best cases.
Every method is drawn by the code that produced its row of Table~\ref{tab:standard},
at the configuration that row reports.
The number in each panel is the mIoU of that image.}
\label{fig:city}
\end{figure*}

Cityscapes tests the same question at a higher resolution. Its $500$ validation
images are $2048\times1024$, and seven of its classes are terms our table already
binds, ``person'', ``car'', ``truck'', ``bus'', ``train'',
``motorcycle'' and ``bicycle''. We score those seven and background, and
ignore ``rider'', which has no term of its own. Every baseline first
reproduces its published number on the standard $19$ classes, within $0.008$ in
every case (Table~\ref{tab:standard}). CAPTOR reaches $0.559$, which is $0.084$ above SCLIP and $0.112$ above ProxyCLIP, each given the
same propagation, with intervals from $+0.052$ to $+0.124$ and from $+0.088$ to
$+0.137$ that resample images. Against Talk2DINO as released, at $0.540$, the difference is $+0.019$ with an interval from $-0.016$ to $+0.049$. Against Talk2DINO without its refinement but given our propagation, at $0.570$, it is $-0.010$ with an interval from $-0.043$ to $+0.019$. CAPTOR is therefore level with
the trained alignment on Cityscapes. Figure~\ref{fig:city} draws four typical
images. CAPTOR is first or second on each, and ahead of both CLIP-based methods
on all four.

\begin{table*}[!tbp]\centering
\caption{Cityscapes validation at $2048\times1024$ under the standard
full-image protocol. Published values come from the paper of each method. The
published and reproduced columns use the standard $19$ classes, and the last two
use the seven classes of the text with background. The last two columns score the
reporting half, first as released and then with the whole-image propagation CAPTOR uses.
Talk2DINO + PAMR is its released configuration, with pixel-adaptive mask
refinement (PAMR). CAPTOR is read at $196$ pixels or at the native
resolution of each image, and its last row adds the background threshold chosen on the
development half, its one choice there.}
\label{tab:standard}
\begin{fitwidth}
\setlength{\tabcolsep}{3.9pt}
\begin{tabular}{lcccc}
\toprule
Method & Published & Reproduced & \makecell{Reporting\\half} & \makecell{Reporting half,\\propagated} \\
\midrule
SCLIP~\cite{sclip} & $0.323$ & $0.324$ & $0.418$ & $0.475$ \\
ProxyCLIP~\cite{proxyclip} & $0.381$ & $0.389$ & $0.414$ & $0.447$ \\
Talk2DINO~\cite{talk2dino} & $0.366$ & $0.367$ & $0.509$ & $\mathbf{0.570}$ \\
Talk2DINO + PAMR~\cite{talk2dino} & $0.381$ & $0.381$ & $0.540$ & $0.538$ \\
CAPTOR, $196$ px & & & $0.372$ & \\
CAPTOR, native & & & $0.404$ & $0.508$ \\
CAPTOR, native, threshold & & & & $\mathbf{0.559}$ \\
\bottomrule
\end{tabular}
\end{fitwidth}
\end{table*}

\subsection{Cross-attention without a text encoder}
\label{sec:attn}

A cross-attention layer first scores every patch against a query and then pools
the image according to those scores. The scores are the compatibility matrix of
Equation~\eqref{eq:attn}, and this section compares the cost of forming them with
a text encoder and with the counted table. For a query term $t$ and an image with
patch tokens $x_1, \dots, x_P$, a standard layer computes
\begin{equation}
a_p = \frac{\exp\big(\langle g(t), h(x_p)\rangle/\sqrt{d}\big)}
 {\sum_{p'} \exp\big(\langle g(t), h(x_{p'})\rangle/\sqrt{d}\big)},
\qquad
o = \sum_p a_p\, h(x_p),
\label{eq:xattn}
\end{equation}
with $g$, $h$ and $d$ as in Section~\ref{sec:m-matrix}. Here $a_p$ is the
attention weight on patch $p$ and $o$ is the pooled output. A new term costs one
text-encoder pass to obtain $g(t)$, which can then be cached, and every image
costs $P$ inner products of dimension $d$ for every term.

CAPTOR replaces the first of these two steps, the term-patch scores, for the terms
its caption corpus supports. The pooling and anything computed after it are left
as they are. Its patch scores are the entries of $A = S\Psi$, and its score for a
whole image is Equation~\eqref{eq:scorefact}, a sum over the $K$ words of the row
of $t$ against the standardized volume vector $z$, which is computed once per
image before any term is known. This replaces only the term-patch scores and does not reproduce the query-dependent pooling of Equation~\eqref{eq:xattn}.

Measured on one GPU (Tables~\ref{tab:cost} to~\ref{tab:termscale},
Section~\ref{sec:s-cost} of the supplement), the methods differ on the term side.
CAPTOR makes a new term queryable by counting, in $15.4$ GFLOPs and $3.2$\,ms,
against $476.8$ GFLOPs and $37.3$\,ms for CLIP's text encoder over its $80$
prompts, and it needs no pretrained text model. Once terms are prepared, scoring an
image against all of them takes every method under a millisecond. Per image the
factorization buys nothing, $2.29$\,ms against Talk2DINO's $2.17$\,ms at $196$
pixels, while Talk2DINO's released configuration, whose refinement cannot be
batched and runs once per term, takes $387$\,ms to label an image with all $78$ terms
against CAPTOR's $8.2$\,ms at a batch of one, a factor of $47$, for $0.023$ more mIoU by Talk2DINO and similar accuracy.

The same table also ranks whole images. Over all $5{,}000$ COCO validation images and the $77$ terms both routes bind, Equation~\eqref{eq:score} reaches $0.947$ mean AUC against $0.903$ for CLIP's pooled image embedding (Section~\ref{sec:s-rank}, Tables~\ref{tab:cocoattn} and~\ref{tab:attnterms}, Figures~\ref{fig:cocoattn}, \ref{fig:querymaps} and~\ref{fig:attnmaps} of the supplement).

Figure~\ref{fig:highres} draws one row of
$A$ at a time, the term-patch scores a cross-attention layer would pool, on ten
photographs of $4$ to $34$ megapixels chosen by a fixed rule, the first
freely licensed Wikimedia Commons quality image of each term, none discarded
after seeing the result. Each is read at $196$ pixels, and every map falls on the
object its term names, including a rusted car in long grass and a dog held by a
child.

\begin{figure*}[!tbp]\centering
\includegraphics[width=\linewidth]{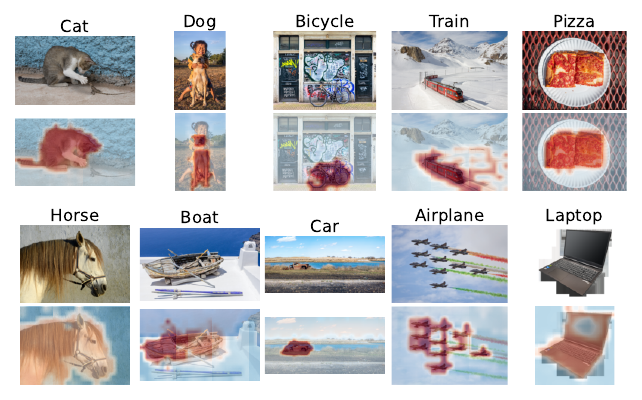}
\caption{One row of the matrix, the score of a single term at every cell, on ten
high-resolution photographs, each above its map. The term of each photograph is
the one it was chosen for. Red is high and blue is low, and each map is
standardized over its own cells, so color shows where the term applies rather
than how strongly. Photographs from Wikimedia
Commons, chosen by the rule the text gives and credited as follows. ``Cat'', ``dog'' and
``bicycle'' are by Basile Morin, ``train'' by Kabelleger and David Gubler,
``pizza'' and ``car'' by Rhododendrites, ``horse'' by Alvesgaspar,
``boat'' by Norbert Nagel, ``airplane'' by Carlos Delgado, all CC BY-SA
$3.0$ or $4.0$, and ``laptop'' by PantheraLeo1359531, CC BY $4.0$.}
\label{fig:highres}
\end{figure*}

\begingroup
\renewenvironment{figure*}[1][]{\begin{figure}[!hbp]}{\end{figure}}
\let\captorig\includegraphics
\renewcommand{\includegraphics}[2][]{\captorig[width=0.85\linewidth]{#2}}
\subsection{Patch-level description in computational pathology}
\label{sec:path}
\label{sec:desc}
\label{sec:term}

CAPTOR is highly useful in histopathology, because an aligned text encoder is
expensive to obtain there and some of the strongest image encoders have none. The
construction is unchanged, with its settings selected on development patients, and
only the image encoder and the caption corpus change.

The vocabulary size is chosen on $41$ of the $103$
BCSS patients and frozen, and the remaining $62$ patients, $134$ fields, are
scored once on the CONCH $28\times28$ grid (Table~\ref{tab:pathmain}). CAPTOR
reaches $0.497$ mIoU. Every baseline reads the CONCH text encoder, is given the
same propagation as CAPTOR, and is offered a per-term standardization that reads
no annotation. The strongest of those that score the CONCH patch features, SCLIP
propagated and standardized, reaches $0.337$. CAPTOR leads it by $+0.160$ with an
interval from $+0.111$ to $+0.206$ that resamples patients. Against dense scoring with CONCH, which a practitioner would otherwise use, without propagation,
the lead is $+0.235$. Figure~\ref{fig:qual} shows a typical field for each
tissue (Tables~\ref{tab:desc2} and~\ref{tab:pathabl} of
Section~\ref{sec:s-path} give each tissue and component). The four vocabularies of CAPTOR, each fitted with its own $k$-means seed,
give $0.429$ to $0.514$ on their own. Averaging them removes most of the
dependence on the seed and is part of the method.

\begin{table}[!tbp]\centering
\caption{Patch-level description of breast tissue on the $134$ fields of the $62$
held-out BCSS patients, five terms, on CONCH's $28\times28$ grid.
Every row reads the CONCH image encoder, and UNI and Phikon are compared in
Table~\ref{tab:encoders}. CAPTOR averages four vocabularies of $K=80$ words, chosen on the other $41$
patients.
The baselines compare the CONCH text embedding of each term with every patch, as
dense scoring~\cite{maskclip} or as SCLIP~\cite{sclip}, and the hybrid compares it with the CAPTOR visual words instead. \emph{Propagated} applies the propagation of
Equation~\eqref{eq:affinity}, and \emph{standardized} rescales each term's scores
over the scored fields without annotation. Intervals resample patients.}
\label{tab:pathmain}
\begin{fitwidth}
\setlength{\tabcolsep}{2.7pt}
\begin{tabular}{lccccc}
\toprule
Method & \makecell{Text\\encoder} & \makecell{Propa-\\gated} & \makecell{Standard-\\ized} & mIoU & \makecell{CAPTOR $-$ this\\ $[95\%$ interval$]$} \\
\midrule
CAPTOR & & $\checkmark$ & & $\mathbf{0.497}$ & \\
CAPTOR, one vocabulary & & $\checkmark$ & & \makecell{$0.429$ to\\ $0.514$} & \\
\midrule
Hybrid & $\checkmark$ & $\checkmark$ & $\checkmark$ & $0.495$ & \makecell{$+0.001$\\ $[-0.024, +0.037]$} \\
SCLIP & $\checkmark$ & $\checkmark$ & $\checkmark$ & $0.337$ & \makecell{$+0.160$\\ $[+0.111, +0.206]$} \\
Dense scoring & $\checkmark$ & $\checkmark$ & $\checkmark$ & $0.329$ & \makecell{$+0.167$\\ $[+0.111, +0.216]$} \\
Dense scoring & $\checkmark$ & $\checkmark$ & & $0.302$ & \makecell{$+0.194$\\ $[+0.131, +0.242]$} \\
Dense scoring & $\checkmark$ & & & $0.262$ & \makecell{$+0.235$\\ $[+0.195, +0.270]$} \\
SCLIP & $\checkmark$ & $\checkmark$ & & $0.257$ & \makecell{$+0.240$\\ $[+0.190, +0.290]$} \\
\bottomrule
\end{tabular}
\end{fitwidth}
\end{table}

The hybrid row of Table~\ref{tab:pathmain} shows where this gain comes from. It
scores the CAPTOR visual words with the CONCH text encoder instead of counted
signatures and reaches $0.495$, level with CAPTOR. The gain over the dense baselines
therefore comes from the visual words and the propagation. Counting and CONCH's text encoder thus name the same visual words with no resolved difference, and counting needs no text encoder. The same holds on all
$200$ fields at $K=40$ without propagation. Two such hybrids reach $0.399$ and
$0.404$ against $0.369$ for CAPTOR, and all three stay well above the $0.309$ of
SCLIP, the strongest baseline that does not use our visual words
(Table~\ref{tab:desc} in the supplement).

Because counting replaces the text
encoder, CAPTOR runs on image encoders that have no text encoder. UNI and Phikon were trained
on slide patches alone, so no text query can be formed against them by any other
method here. Table~\ref{tab:encoders} reads all three encoders on the $14\times14$
grid they share, each with its own vocabularies counted from the same captions and
its settings chosen on development patients, and compares each with the strongest
reading of CONCH's text encoder chosen the same way. On the $62$ held-out breast
patients, UNI read through CAPTOR reaches $0.456$ mIoU against $0.324$, a lead of
$+0.132$ with an interval from $+0.077$ to $+0.185$. An encoder with no language
of its own, given only captions, thus names breast-cancer tissue better than a text
encoder trained for pathology. Phikon is level, at $-0.025$ with an interval
containing zero. Both use $K=80$, chosen on development patients. At $K=40$, Phikon left most stroma in one word that every caption term scores near zero, so
no name could be counted for it.

\begin{table}[!tbp]\centering
\caption{Three image encoders read through CAPTOR on two organs, on the
$14\times14$ grid all three produce. The settings of each family, per encoder for CAPTOR and for the baselines, are chosen on development patients ($41$ of $103$ for BCSS,
$10$ of $25$ for LUAD-HistoSeg) and read once on the rest. The baseline is the
strongest reading of CONCH's own text encoder so chosen, given CAPTOR's
propagation and a per-term standardization that reads no annotation. Only CONCH has a text encoder of its own. Differences are CAPTOR minus that baseline, with $95\%$ intervals that resample patients.}
\label{tab:encoders}
\begin{fitwidth}
\begin{tabular}{lcccc}
\toprule
 & \multicolumn{2}{c}{breast, BCSS} & \multicolumn{2}{c}{lung, LUAD-HistoSeg} \\
CAPTOR on & mIoU & \makecell{Minus\\ baseline} & mIoU & \makecell{Minus\\ baseline} \\
\midrule
CONCH & $\mathbf{0.521}$ & \makecell{$+0.197$\\ $[+0.140, +0.255]$} & $0.325$ & \makecell{$-0.059$\\ $[-0.113, -0.005]$} \\
UNI & $0.456$ & \makecell{$+0.132$\\ $[+0.077, +0.185]$} & $\mathbf{0.410}$ & \makecell{$+0.026$\\ $[-0.056, +0.104]$} \\
Phikon & $0.299$ & \makecell{$-0.025$\\ $[-0.074, +0.021]$} & $0.374$ & \makecell{$-0.010$\\ $[-0.051, +0.033]$} \\
\midrule
Baseline, CONCH text & $0.324$ & & $0.384$ & \\
\bottomrule
\end{tabular}
\end{fitwidth}
\end{table}

On a second organ, lung-cancer tissue scored cell by cell against the pixel annotation
of LUAD-HistoSeg, CAPTOR on UNI and on Phikon is level with the baseline that reads the CONCH text encoder, while CAPTOR on CONCH falls behind it by $0.059$
(Table~\ref{tab:enctissue} and Figure~\ref{fig:luadqual} in the supplement give each tissue). The table is
the same one counted from PathCap captions, and only its settings are chosen on
lung development patients. The loss comes from one tissue. CAPTOR gives $48\%$ of
tumor-associated stroma cells to ``tumor'', and still $42\%$ of the cells whose
annotation is at least $95\%$ stroma, so the error is not confined to tissue
borders. The visual words themselves separate the two tissues, since naming each
word after the tissue it mostly holds would reach $0.475$ mIoU on CONCH's own grid.
What fails is the names. Pathologists caption the whole lesion as ``tumor'', its
stroma included, whereas this annotation separates the two, and a signature
counted from captions takes on the meaning of the captions.  A wider test that names visual words also finds no resolved difference. Across six pathology encoders and three
benchmarks, CAPTOR is ahead of a text encoder adapted on the same captions in $12$
of $18$ combinations, but level on average, at $-0.002$ AUC (Tables~\ref{tab:nct} and~\ref{tab:naming6} in the
supplement).

A new terminology starts without an aligned
text encoder. On all $200$ fields, dense scoring with a
general-purpose text encoder reaches $0.084$ mIoU, the domain text
encoder CONCH $0.328$, and CAPTOR with a single vocabulary $0.465$. Re-estimated from subsets of the corpus and without the
propagation, CAPTOR reaches $0.287$ from $649$ captions, $0.340$ from $1{,}623$ and $0.396$ from all $32{,}463$. Even the smallest subset exceeds the CONCH text encoder measured the same way, at $0.276$, from far fewer captions than an aligned encoder is trained on (Table~\ref{tab:pathextra} in the supplement). 

\begin{figure*}[!tbp]\centering
\includegraphics[width=\linewidth]{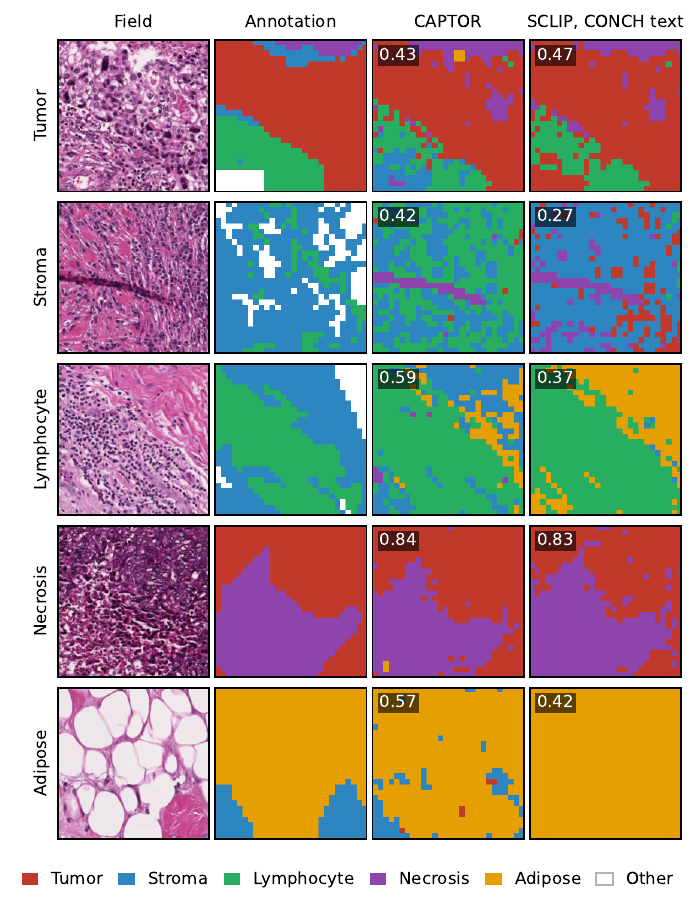}
\caption{Named segmentation of breast fields with $K=80$, four vocabularies and the
propagation. The columns are the field, its annotation, the term CAPTOR scores
highest, and SCLIP reading the CONCH text encoder with the same propagation and
standardization, the strongest baseline of Table~\ref{tab:pathmain} that does not
borrow the words of CAPTOR. Each row shows the field whose intersection over union
(IoU) for its tissue is the median among the fields that tissue dominates, so the
panel shows the typical case. Maps are drawn on the $28\times28$ scoring grid, and
each number is that field's mIoU. CAPTOR is ahead on four of the five fields.}
\label{fig:qual}
\end{figure*}

\FloatBarrier
\endgroup
\section{Discussion}
\label{sec:limits}

The CAPTOR vocabulary is the set of terms
a caption corpus mentions often enough to clear their nulls. Objects and tissues
are named in captions and bind well, while surfaces and materials are rarely
named, so scene-parsing benchmarks built on them lie outside what the method addresses.
The binding also inherits the level of detail of the captions, as lung stroma shows (Section~\ref{sec:path}). An object photographed almost always in one setting absorbs the words of that setting, which is where CAPTOR loses
on COCO-Object. Attributes, relations and negation were not tested.

A second limit lies in the partition. Patches that share a visual word are
identical to every term under the hard assignment, so CAPTOR cannot separate what the encoder partition does not separate. The soft assignment (Table~\ref{tab:loc2} in the
supplement), the propagation and
joint words relax this, and averaging vocabularies fitted with different seeds
removes most of the dependence on any one partition. 

Captions supervise the table and development
labels choose a few settings (Section~\ref{sec:setup}), so CAPTOR is not free of
supervision, and neither are the text encoders its baselines read. What it removes is gradient training on the pairs and the text encoder.

\section{Conclusion}

We introduced CAPTOR, which estimates the matrix that scores every term against
every patch from captions over the visual words of a frozen encoder, where a vision-language model needs a text encoder to form it.   On natural images, CAPTOR leads the training-free baselines on COCO, Pascal VOC and Cityscapes, and it is
level there with a text alignment trained on the same backbone. In pathology, CAPTOR gives a text interface to encoders that have
none, from captions alone and without gradients. Through it, an encoder
trained on slides alone names breast tissue better than dense scoring with a trained pathology text encoder. Grounding text in images therefore does not require a text encoder, only captioned images and counting. This opens grounding to many specialized domains in which aligned text encoders and dense annotations are unavailable or costly to obtain.

\section*{CRediT authorship contribution statement}
\textbf{Elad Yoshai:} Conceptualization, Methodology, Software, Validation,
Formal analysis, Investigation, Data curation, Writing of the original draft,
Visualization. \textbf{Natan T. Shaked:} Conceptualization, Supervision, Writing of the original draft, Writing, review and editing, Project administration.

\section*{Declaration of competing interest}
The authors declare that they have no known competing financial interests or
personal relationships that could have appeared to influence the work reported
in this paper.

\section*{Data and code availability}
All data are public and de-identified, so no institutional review board approval
was required and no new patient data were collected. PathCap is distributed with
PathAsst~\cite{pathcap}, and BCSS~\cite{bcss}, LUAD-HistoSeg~\cite{luadhistoseg},
COCO~\cite{coco}, Pascal VOC~\cite{voc} and Cityscapes~\cite{cityscapes} are
obtained from their published releases. Code to reproduce every number will be released publicly upon acceptance.

\section*{Declaration of generative AI and AI-assisted technologies in the writing process}
During the preparation of this work the authors used Claude (Anthropic) to assist
with editing the text, checking reported numbers against the experimental
records, and writing analysis code. After using this tool, the authors reviewed
and edited the content as needed and take full responsibility for the content of
the publication.

\section*{Acknowledgements}
This work was funded by a digital pathology grant from the Ministry of Science and Technology of Israel. We thank the
maintainers of the public datasets used here and the authors of CONCH, UNI and
Phikon for releasing model weights.

\bibliographystyle{elsevier/elsarticle-num-etal3}
\bibliography{refs}

\begin{thebibliography}{10}
\expandafter\ifx\csname url\endcsname\relax
  \def\url#1{\texttt{#1}}\fi
\expandafter\ifx\csname urlprefix\endcsname\relax\def\urlprefix{URL }\fi
\expandafter\ifx\csname href\endcsname\relax
  \def\href#1#2{#2} \def\path#1{#1}\fi

\bibitem{conch}
M.~Y. Lu, B.~Chen, D.~F.~K. Williamson, et~al., A visual-language foundation
  model for computational pathology, Nature Medicine 30~(3) (2024) 863-874.
\newblock \href {https://doi.org/10.1038/s41591-024-02856-4}
  {\path{doi:10.1038/s41591-024-02856-4}}.

\bibitem{uni}
R.~J. Chen, T.~Ding, M.~Y. Lu, et~al., Towards a general-purpose foundation
  model for computational pathology, Nature Medicine 30~(3) (2024) 850-862.
\newblock \href {https://doi.org/10.1038/s41591-024-02857-3}
  {\path{doi:10.1038/s41591-024-02857-3}}.

\bibitem{virchow}
E.~Vorontsov, A.~Bozkurt, A.~Casson, et~al., A foundation model for
  clinical-grade computational pathology and rare cancers detection, Nature
  Medicine 30~(10) (2024) 2924-2935.
\newblock \href {https://doi.org/10.1038/s41591-024-03141-0}
  {\path{doi:10.1038/s41591-024-03141-0}}.

\bibitem{gigapath}
H.~Xu, N.~Usuyama, J.~Bagga, et~al., A whole-slide foundation model for digital
  pathology from real-world data, Nature 630~(8015) (2024) 181-188.
\newblock \href {https://doi.org/10.1038/s41586-024-07441-w}
  {\path{doi:10.1038/s41586-024-07441-w}}.

\bibitem{talk2dino}
L.~Barsellotti, L.~Bianchi, N.~Messina, et~al., Talking to {DINO}: Bridging
  self-supervised vision backbones with language for open-vocabulary
  segmentation, in: Proceedings of the IEEE/CVF International Conference on
  Computer Vision, 2025, pp. 22025-22035.
\newblock \href {https://doi.org/10.1109/ICCV51701.2025.02045}
  {\path{doi:10.1109/ICCV51701.2025.02045}}.

\bibitem{mori1999}
Y.~Mori, H.~Takahashi, R.~Oka, Image-to-word transformation based on dividing
  and vector quantizing images with words, in: First International Workshop on
  Multimedia Intelligent Storage and Retrieval Management, 1999.

\bibitem{duygulu}
P.~Duygulu, K.~Barnard, N.~de~Freitas, et~al., Object recognition as machine
  translation: Learning a lexicon for a fixed image vocabulary, in: Proceedings
  of the European Conference on Computer Vision, Vol. 2353 of LNCS, 2002, pp.
  97-112.
\newblock \href {https://doi.org/10.1007/3-540-47979-1_7}
  {\path{doi:10.1007/3-540-47979-1_7}}.

\bibitem{barnard}
K.~Barnard, P.~Duygulu, D.~Forsyth, et~al., Matching words and pictures,
  Journal of Machine Learning Research 3 (2003) 1107-1135.

\bibitem{clip}
A.~Radford, J.~W. Kim, C.~Hallacy, et~al., Learning transferable visual models
  from natural language supervision, in: Proceedings of the International
  Conference on Machine Learning, Vol. 139 of PMLR, 2021, pp. 8748-8763.

\bibitem{coco}
T.-Y. Lin, M.~Maire, S.~Belongie, et~al., Microsoft {COCO}: Common objects in
  context, in: Proceedings of the European Conference on Computer Vision, LNCS,
  2014, pp. 740-755.
\newblock \href {https://doi.org/10.1007/978-3-319-10602-1_48}
  {\path{doi:10.1007/978-3-319-10602-1_48}}.

\bibitem{voc}
M.~Everingham, S.~M.~A. Eslami, L.~Van~Gool, et~al., The {PASCAL} visual object
  classes challenge: A retrospective, International Journal of Computer Vision
  111~(1) (2015) 98-136.
\newblock \href {https://doi.org/10.1007/s11263-014-0733-5}
  {\path{doi:10.1007/s11263-014-0733-5}}.

\bibitem{cityscapes}
M.~Cordts, M.~Omran, S.~Ramos, et~al., The {Cityscapes} dataset for semantic
  urban scene understanding, in: Proceedings of the IEEE/CVF Conference on
  Computer Vision and Pattern Recognition, 2016, pp. 3213-3223.
\newblock \href {https://doi.org/10.1109/CVPR.2016.350}
  {\path{doi:10.1109/CVPR.2016.350}}.

\bibitem{plip}
Z.~Huang, F.~Bianchi, M.~Yuksekgonul, et~al., A visual-language foundation
  model for pathology image analysis using medical {Twitter}, Nature Medicine
  29~(9) (2023) 2307-2316.
\newblock \href {https://doi.org/10.1038/s41591-023-02504-3}
  {\path{doi:10.1038/s41591-023-02504-3}}.

\bibitem{maskclip}
C.~Zhou, C.~C. Loy, B.~Dai, Extract free dense labels from {CLIP}, in:
  Proceedings of the European Conference on Computer Vision, LNCS, 2022, pp.
  696-712.
\newblock \href {https://doi.org/10.1007/978-3-031-19815-1_40}
  {\path{doi:10.1007/978-3-031-19815-1_40}}.

\bibitem{sclip}
F.~Wang, J.~Mei, A.~Yuille, {SCLIP}: Rethinking self-attention for dense
  vision-language inference, in: Proceedings of the European Conference on
  Computer Vision, LNCS, 2024, pp. 315-332.
\newblock \href {https://doi.org/10.1007/978-3-031-72664-4_18}
  {\path{doi:10.1007/978-3-031-72664-4_18}}.

\bibitem{proxyclip}
M.~Lan, C.~Chen, Y.~Ke, et~al., {ProxyCLIP}: Proxy attention improves {CLIP}
  for open-vocabulary segmentation, in: Proceedings of the European Conference
  on Computer Vision, LNCS, 2024, pp. 70-88.
\newblock \href {https://doi.org/10.1007/978-3-031-73113-6_5}
  {\path{doi:10.1007/978-3-031-73113-6_5}}.

\bibitem{dinov2}
M.~Oquab, T.~Darcet, T.~Moutakanni, et~al., {DINOv2}: Learning robust visual
  features without supervision, Transactions on Machine Learning Research
  (2024).

\bibitem{dinotxt}
C.~Jose, T.~Moutakanni, D.~Kang, et~al., {DINOv2} meets text: A unified
  framework for image- and pixel-level vision-language alignment, in:
  Proceedings of the IEEE/CVF Conference on Computer Vision and Pattern
  Recognition, 2025, pp. 24905-24916.
\newblock \href {https://doi.org/10.1109/CVPR52734.2025.02319}
  {\path{doi:10.1109/CVPR52734.2025.02319}}.

\bibitem{sam3}
N.~Carion, L.~Gustafson, Y.-T. Hu, et~al., {SAM} 3: Segment anything with
  concepts, in: International Conference on Learning Representations, 2026.

\bibitem{zoomshot}
J.~Shipard, A.~Wiliem, K.~N. Thanh, et~al., Zoom-shot: Fast, efficient and
  unsupervised zero-shot knowledge transfer from {CLIP} to vision encoders,
  Pattern Recognition 172 (2026) 112323.
\newblock \href {https://doi.org/10.1016/j.patcog.2025.112323}
  {\path{doi:10.1016/j.patcog.2025.112323}}.

\bibitem{phikon}
A.~Filiot, P.~Jacob, A.~Mac~Kain, et~al., {Phikon-v2}, a large and public
  feature extractor for biomarker prediction, arXiv preprint arXiv:2409.09173
  (2024).
\newblock \href {https://doi.org/10.48550/arXiv.2409.09173}
  {\path{doi:10.48550/arXiv.2409.09173}}.

\bibitem{mahmood_survey}
A.~H. Song, G.~Jaume, D.~F.~K. Williamson, et~al., Artificial intelligence for
  digital and computational pathology, Nature Reviews Bioengineering 1~(12)
  (2023) 930-949.
\newblock \href {https://doi.org/10.1038/s44222-023-00096-8}
  {\path{doi:10.1038/s44222-023-00096-8}}.

\bibitem{vdpath}
R.~Huang, H.~Lin, W.~Zhuo, et~al., Fine-grained alignment in medical pathology
  vision-language models via variational distillation, Pattern Recognition 180
  (2026) 114283.
\newblock \href {https://doi.org/10.1016/j.patcog.2026.114283}
  {\path{doi:10.1016/j.patcog.2026.114283}}.

\bibitem{jeon2003}
J.~Jeon, V.~Lavrenko, R.~Manmatha, Automatic image annotation and retrieval
  using cross-media relevance models, in: Proceedings of the ACM SIGIR
  Conference on Research and Development in Information Retrieval, 2003, pp.
  119-126.
\newblock \href {https://doi.org/10.1145/860435.860459}
  {\path{doi:10.1145/860435.860459}}.

\bibitem{vit}
A.~Dosovitskiy, L.~Beyer, A.~Kolesnikov, et~al., An image is worth {16x16}
  words: Transformers for image recognition at scale, in: International
  Conference on Learning Representations, 2021.

\bibitem{kmeanspp}
D.~Arthur, S.~Vassilvitskii, k-means++: The advantages of careful seeding, in:
  Proceedings of the ACM-SIAM Symposium on Discrete Algorithms, 2007, pp.
  1027-1035.

\bibitem{pathcap}
Y.~Sun, C.~Zhu, S.~Zheng, et~al., {PathAsst}: A generative foundation {AI}
  assistant towards artificial general intelligence of pathology, in:
  Proceedings of the AAAI Conference on Artificial Intelligence, Vol.~38, 2024,
  pp. 5034-5042.
\newblock \href {https://doi.org/10.1609/aaai.v38i5.28308}
  {\path{doi:10.1609/aaai.v38i5.28308}}.

\bibitem{bcss}
M.~Amgad, H.~Elfandy, H.~Hussein, et~al., Structured crowdsourcing enables
  convolutional segmentation of histology images, Bioinformatics 35~(18) (2019)
  3461-3467.
\newblock \href {https://doi.org/10.1093/bioinformatics/btz083}
  {\path{doi:10.1093/bioinformatics/btz083}}.

\bibitem{luadhistoseg}
C.~Han, J.~Lin, J.~Mai, et~al., Multi-layer pseudo-supervision for
  histopathology tissue semantic segmentation using patch-level classification
  labels, Medical Image Analysis 80 (2022) 102487.
\newblock \href {https://doi.org/10.1016/j.media.2022.102487}
  {\path{doi:10.1016/j.media.2022.102487}}.

\bibitem{registers}
T.~Darcet, M.~Oquab, J.~Mairal, et~al., Vision transformers need registers, in:
  International Conference on Learning Representations, 2024.

\bibitem{clipdinoiser}
M.~Wysocza{\'n}ska, O.~Sim{\'e}oni, M.~Ramamonjisoa, et~al., {CLIP-DINOiser}:
  Teaching {CLIP} a few {DINO} tricks for open-vocabulary semantic
  segmentation, in: Proceedings of the European Conference on Computer Vision,
  LNCS, 2024, pp. 320-337.
\newblock \href {https://doi.org/10.1007/978-3-031-73030-6_18}
  {\path{doi:10.1007/978-3-031-73030-6_18}}.

\bibitem{wordnet}
G.~A. Miller, {WordNet}: A lexical database for {English}, Communications of
  the ACM 38~(11) (1995) 39-41.
\newblock \href {https://doi.org/10.1145/219717.219748}
  {\path{doi:10.1145/219717.219748}}.

\bibitem{efron}
B.~Efron, Bootstrap methods: Another look at the jackknife, The Annals of
  Statistics 7~(1) (1979) 1-26.
\newblock \href {https://doi.org/10.1214/aos/1176344552}
  {\path{doi:10.1214/aos/1176344552}}.

\bibitem{lora}
E.~J. Hu, Y.~Shen, P.~Wallis, et~al., {LoRA}: Low-rank adaptation of large
  language models, in: International Conference on Learning Representations,
  2022.

\bibitem{pannuke}
J.~Gamper, N.~Alemi~Koohbanani, K.~Benes, et~al., {PanNuke} dataset extension,
  insights and baselines, arXiv preprint arXiv:2003.10778 (2020).
\newblock \href {https://doi.org/10.48550/arXiv.2003.10778}
  {\path{doi:10.48550/arXiv.2003.10778}}.

\bibitem{nct}
J.~N. Kather, J.~Krisam, P.~Charoentong, et~al., Predicting survival from
  colorectal cancer histology slides using deep learning: A retrospective
  multicenter study, PLOS Medicine 16~(1) (2019) e1002730.
\newblock \href {https://doi.org/10.1371/journal.pmed.1002730}
  {\path{doi:10.1371/journal.pmed.1002730}}.

\bibitem{quilt}
W.~O. Ikezogwo, M.~S. Seyfioglu, F.~Ghezloo, et~al., {Quilt-1M}: One million
  image-text pairs for histopathology, in: Advances in Neural Information
  Processing Systems, Vol.~36, 2023, pp. 37995-38017.

\bibitem{biomedclip}
S.~Zhang, Y.~Xu, N.~Usuyama, et~al., A multimodal biomedical foundation model
  trained from fifteen million image-text pairs, NEJM AI 2~(1) (2025)
  AIoa2400640.
\newblock \href {https://doi.org/10.1056/AIoa2400640}
  {\path{doi:10.1056/AIoa2400640}}.

\end{thebibliography}

\let\origappendix\appendix
\renewcommand{\appendix}{\origappendix
  \setcounter{section}{0}\setcounter{figure}{0}\setcounter{table}{0}
  \renewcommand{\thesection}{S\arabic{section}}%
  \renewcommand{\thefigure}{S\arabic{figure}}%
  \renewcommand{\thetable}{S\arabic{table}}%
  \section*{Supplementary material}}
\clearpage
\onecolumn
\appendix

This supplement follows the order of the paper, and each table and figure can be
read on its own from its caption.

The abbreviations follow the main text. CAPTOR is the Caption-Anchored
Patch-Term Occurrence Representation. mIoU is the mean intersection over union
(IoU) over classes, and AUC the area under the receiver operating characteristic
curve. CLIP is Contrastive Language-Image Pre-training, COCO is Common Objects in
Context, VOC is the Pascal Visual Object Classes benchmark, and BCSS is the Breast
Cancer Semantic Segmentation dataset. LoRA is low-rank adaptation, SAM~3 is
Segment Anything Model~3, PAMR is pixel-adaptive mask refinement, and CPU and GPU
are the central and graphics processing units.

\section{Datasets, protocols and settings}
\label{sec:s-setup}

The matched protocol scores only grid cells that one object class covers by more
than half. Captions rarely name background, so scoring it would compare rejection
thresholds rather than how methods score terms. The rule is the same for every
method, so comparisons are valid, but its values are not comparable with published
mIoU, which includes background. Every method reads the same single center crop,
whereas the standard protocols run each baseline with its own released settings.

Terms are matched to captions through their WordNet~\cite{wordnet} forms. When one
term's name contains another, as ``teddy bear'' contains ``bear'', plain matching
gives the two nearly the same signature. Two rules prevent this. A phrase is
matched before its parts, and a word form shared with another term is dropped.
Terms whose bare word is too general, such as ``stop sign'', require their full
name, and the development images select eight such terms (Table~\ref{tab:hyper}).
Neither rule changes anything on VOC or in pathology.

\begin{figure*}[!htp]\centering
\includegraphics[width=0.66\linewidth]{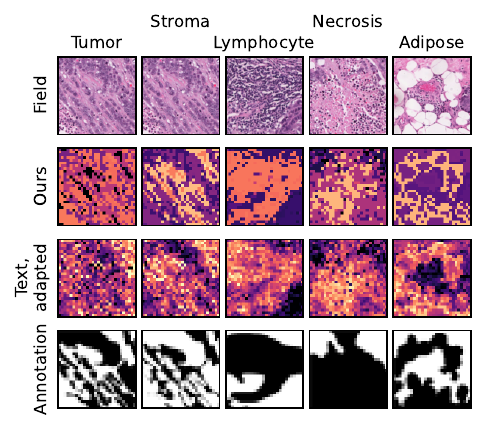}
\caption{Score maps for one tissue term at a time on BCSS fields. Each field is the first
in which the queried tissue covers more than a third of the annotation, a rule fixed
before drawing. Row two is CAPTOR's map, which looks up the term's signature at each
patch's visual word. Row three is the same term read by the adapted CONCH encoder, and
row four is the annotation for that term. All maps are rescaled to a common
range.}
\label{fig:attnmaps}
\end{figure*}

Two baselines are trained on the same $32{,}463$ PathCap caption pairs that CAPTOR
counts. Each tunes a text encoder against CONCH's frozen image encoder with a
contrastive loss, training only rank-$16$ low-rank adapters (LoRA)~\cite{lora} for
six epochs. The \emph{adapted CONCH encoder} tunes CONCH's own text encoder, with
$1{,}179{,}648$ trained parameters, and is the baseline for ranking fields
(Table~\ref{tab:attnterms}, Figure~\ref{fig:attnmaps}). The \emph{adapted CLIP
encoder} tunes the text encoder of CLIP ViT-B/32~\cite{clip} and adds a linear map
into CONCH's image space, with $1{,}049{,}088$ trained parameters. Its pairs are
split by source article, and it is the baseline for naming visual words (Table~\ref{tab:naming6}).

Intervals are percentile bootstraps over images for natural images and over
patients in pathology. NCT-CRC has no patient identifier, so its intervals resample tiles and may understate the uncertainty that patient grouping would add.

A journal article can reproduce a crop of a public cohort that an evaluation set
also draws on. We therefore compared every evaluation image with all $110{,}000$
PathCap figures by the cosine similarity of their CONCH image embeddings
(Table~\ref{tab:overlap}). No pair reaches $0.95$, and the $24$ closest pairs show
different images of similar tissue. A small crop inside a multi-panel figure would
escape this test.

\begin{table*}[!htp]\centering
\caption{Every dataset, what it is used for, and how it is split. Development
images or patients choose settings, and reporting images or patients are scored
once. A patient-level split keeps all images of one patient on one side. No
evaluation image enters the table of signatures, which is counted from the
caption corpus alone.}
\label{tab:datasets}
\begin{fitwidth}
\setlength{\tabcolsep}{3pt}
\begin{tabular}{>{\raggedright\arraybackslash\hyphenpenalty=10000}p{3.0cm}>{\raggedright\arraybackslash\hyphenpenalty=10000}p{3.2cm}>{\raggedright\arraybackslash\hyphenpenalty=10000}p{2.3cm}>{\raggedright\arraybackslash\hyphenpenalty=10000}p{2.5cm}}
\toprule
Dataset & Use & Images & \makecell[l]{development /\\ reporting} \\
\midrule
\multicolumn{4}{l}{\emph{caption corpora}} \\
COCO training captions~\cite{coco} & Table for natural images & $15{,}000$, or $30{,}000$ for VOC & Disjoint from evaluation \\
PathCap~\cite{pathcap} & Table for pathology & $32{,}463$ figures & Disjoint from evaluation \\
\midrule
\multicolumn{4}{l}{\emph{natural images}} \\
COCO validation & Matched protocol & $5{,}000$ & $2{,}500$ / $2{,}500$ images \\
Pascal VOC 2012 val.~\cite{voc} & Transfer from COCO & $1{,}449$ & $724$ / $725$ images \\
COCO-Object~\cite{sclip} & Standard protocol & $5{,}000$ & $2{,}500$ / $2{,}500$ images \\
Cityscapes val.~\cite{cityscapes} & Standard protocol & $500$ & $250$ / $250$ images \\
\midrule
\multicolumn{4}{l}{\emph{pathology}} \\
BCSS~\cite{bcss} & Segmentation & $200$ fields & $41$ / $62$ patients \\
LUAD-HistoSeg~\cite{luadhistoseg} & Segmentation & $607$ tiles & $10$ / $15$ patients \\
LUAD-HistoSeg & Naming and field ranking & $564$ tiles & Read once \\
PanNuke~\cite{pannuke} & Naming and field ranking & $6{,}785$ tiles, $19$ organs & Read once \\
NCT-CRC~\cite{nct} & Naming in a further organ & $1{,}980$ tiles, $220$ per class & Read once \\
\bottomrule
\end{tabular}
\end{fitwidth}
\end{table*}

\begin{table*}[!htp]\centering
\caption{Every constant, and where each was chosen. Values selected on
development data are marked, and the rest are fixed once and never swept. The
pathology column is the configuration Section~\ref{sec:path} reports, in which the
selection sets $\lambda$ large and runs no recounting, which by
Equation~\eqref{eq:trust} is the averaged scoring rule. The development images select eight natural-image terms whose full name is required
(Section~\ref{sec:setup}). They are ``traffic light'', ``stop sign'',
``parking meter'', ``baseball bat'', ``baseball glove'', ``tennis
racket'', ``wine glass'' and ``hot dog''.}
\label{tab:hyper}
\begin{fitwidth}
\begin{tabular}{lp{3.3cm}ccl}
\toprule
Symbol & Meaning & Natural images & Pathology & Chosen \\
\midrule
$K$ & Words per vocabulary & $640$ & $80$ & Development \\
$J$ & Vocabularies averaged & $4$ & $4$ & Fixed \\
$m$ & Words a patch carries, Eq.~\eqref{eq:softscore} & $16$ & $1$ & Development \\
$\beta$ & Sharpness of those weights & $10$ & $10$ & Development \\
$\lambda$ & Captions for half belief, Eq.~\eqref{eq:trust} & $3$ & $3\times10^{4}$ & Development \\
$\varepsilon$ & Responsibility floor, Eq.~\eqref{eq:em} & $0.1$ & $0.1$ & Fixed \\
$\tau$ & Responsibility sharpness, Eq.~\eqref{eq:em} & $1$ & Not used & Development \\
 & Recounting rounds & $1$ & $0$ & Development \\
$\theta$, $\kappa$ & Affinity threshold and gain, Eq.~\eqref{eq:affinity} & $1.2$, $3$ & $1.2$, $3$ & Fixed \\
 & Division by the null, Eq.~\eqref{eq:nullscale} & On & Off & Development \\
 & Permutation draws for $\sigma_t$ & $200$ & $200$ & Fixed \\
 & Null percentile & $95$ & $95$ & Fixed \\
 & Development share, split seed & $0.5$, $0$ & $0.4$, $0$ & Fixed \\
\bottomrule
\end{tabular}
\end{fitwidth}
\end{table*}

\begin{table}[!htp]\centering
\caption{Nearest PathCap figure for every evaluation image, by the cosine
similarity of CONCH image embeddings. No evaluation image comes close to a
duplicate, which would score near $1$. One LUAD-HistoSeg archive entry is a
file-system metadata file rather than an image and is excluded.}
\label{tab:overlap}
\begin{fitwidth}
\begin{tabular}{lrcccc}
\toprule
Evaluation set & Images & Median & 99th percentile & Maximum & Above $0.95$ \\
\midrule
BCSS regions & $151$ & $0.757$ & $0.866$ & $0.882$ & $0$ \\
BCSS tiles & $6{,}736$ & $0.776$ & $0.866$ & $0.904$ & $0$ \\
LUAD-HistoSeg & $607$ & $0.734$ & $0.850$ & $0.868$ & $0$ \\
PanNuke & $6{,}785$ & $0.769$ & $0.860$ & $0.904$ & $0$ \\
NCT-CRC & $7{,}180$ & $0.737$ & $0.853$ & $0.909$ & $0$ \\
\bottomrule
\end{tabular}
\end{fitwidth}
\end{table}

\FloatBarrier

\section{The method on real images}
\label{sec:s-method}

\begin{figure*}[!htp]\centering
\includegraphics[width=0.9\linewidth]{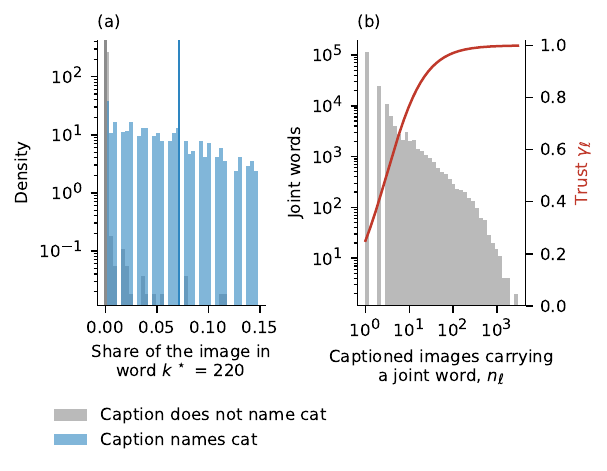}
\caption{Two further steps of the method on COCO, from one vocabulary of the
reported table. \textbf{(a)} The contrast Equation~\eqref{eq:signature} counts,
for ``cat'' and the word its signature is largest on: the share of each image in
that word, for images whose captions name ``cat'' and for the rest, with each
group's mean marked. Images whose captions name ``cat'' give that word a mean
share of $0.071$, against $0.00006$ for the rest. \textbf{(b)} Why the score of
Equation~\eqref{eq:trust} blends two estimates. A joint word combines one word from
each of the four vocabularies. The bars count joint words by the number of
captioned images that contain them, and the red curve is the trust weight
$\gamma_\ell$ at $\lambda = 3$. Of the $178{,}828$ joint words, $83\%$ appear in
three images or fewer, where $\gamma_\ell$ is at most one half. Most patches
therefore rely mainly on the better-supported average of the four vocabularies,
and only frequent joint words are trusted on their own.}
\label{fig:mechsupp}
\end{figure*}

Figure~\ref{fig:mechsupp} draws two steps of Section~\ref{sec:method} that
Figure~\ref{fig:mechanism} leaves out, on COCO and from one vocabulary of the
reported table.

\FloatBarrier

\section{COCO under the matched protocol}
\label{sec:s-coco}
\label{sec:supp-coco}

This section gives the qualitative results on COCO, every ablation and the comparison
of estimators. All tables use the COCO validation images and
the table counted on the disjoint training captions.

The margin over ProxyCLIP given the same smoothing grows with $K$
(Table~\ref{tab:coco-k}), from $-0.016$ at $K=320$, the one size at which ProxyCLIP
leads, to $+0.055$ at $K=2560$. From $K=640$ upward its interval excludes zero.
Given the same smoothing, ProxyCLIP scores $0.506$ on all $5{,}000$ images at every $K$.

\begin{figure*}[!htp]\centering
\includegraphics[width=0.71\linewidth]{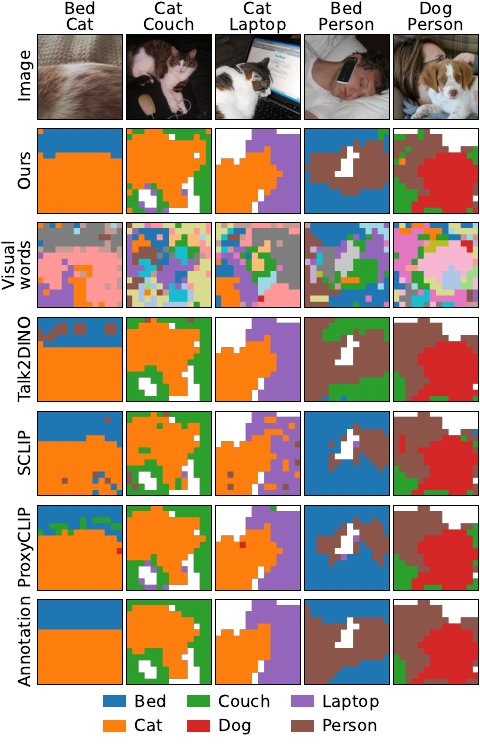}
\caption{COCO as a named segmentation, at the reported configuration. Each column is one
validation image, chosen from the annotation alone and never from any method's
output. It contains exactly two of the terms shown, each covering at least $35\%$
of the center crop and together at least $70\%$, and no pair of terms appears
twice. Row two is the term CAPTOR scores highest, and row three the same patches' visual
words with no names attached. Rows four to six are Talk2DINO, SCLIP and ProxyCLIP from
their authors' released code, and row seven is the annotation.}
\label{fig:coco}
\end{figure*}

\begin{table}[!htp]\centering
\caption{Components of the natural-image method at $K=1280$, on $5{,}000$ COCO
validation images with the table estimated on the disjoint caption corpus.
Captions here are read by the plain word-form rule rather than the phrase-aware
reading of Section~\ref{sec:setup}. These values are therefore comparable within this
table and lower than those of Table~\ref{tab:coco-k}. \emph{Registers} uses
DINOv2 with register tokens~\cite{registers}, extra tokens that absorb image-wide
information that plain DINOv2 stores in a few background patches, so the patch
features are cleaner. \emph{Propagation} is $\Psi = \Phi W^{\top}$ of
Equation~\eqref{eq:affinity}. \emph{Synonyms} also counts a caption as naming a
term when it uses one of the term's WordNet forms, such as ``sofa'' for
``couch''. $\Delta$
is the change from the first row. The last row removes the division by the
permutation null of Equation~\eqref{eq:nullscale} from the configuration above
it, which leaves every term's ranking of patches unchanged.}
\label{tab:coco-abl}
\begin{fitwidth}
\begin{tabular}{lcccc}
\toprule
Configuration & mIoU & $\Delta$ & Acc. & Term AUC \\
\midrule
Base & $0.400$ & & $0.569$ & $0.962$ \\
$+$ synonyms & $0.393$ & $-0.008$ & $0.627$ & $0.961$ \\
$+$ registers & $0.458$ & $+0.058$ & $0.597$ & $0.967$ \\
$+$ propagation & $0.483$ & $+0.083$ & $0.638$ & $0.981$ \\
$+$ propagation $+$ synonyms & $0.468$ & $+0.068$ & $0.696$ & $0.980$ \\
$+$ registers $+$ synonyms & $0.459$ & $+0.059$ & $0.658$ & $0.966$ \\
$+$ registers $+$ propagation & $0.510$ & $+0.110$ & $0.643$ & $0.984$ \\
All three & $\mathbf{0.511}$ & $+0.110$ & $0.713$ & $0.983$ \\
\midrule
All three, without null scaling & $0.429$ & $+0.028$ & $0.483$ & $0.983$ \\
\bottomrule
\end{tabular}
\end{fitwidth}
\end{table}

\begin{table}[!htp]\centering
\caption{How the vocabulary size $K$ changes CAPTOR on COCO, with all $5{,}000$
evaluation images scored. \emph{mIoU} is CAPTOR's score. \emph{CAPTOR $-$ ProxyCLIP} subtracts the mIoU of
ProxyCLIP given the same smoothing, which scores $0.506$ on these images at every
$K$. The \emph{label-free score} reads no
annotation. Signatures counted on half of the captions rank the other half's images
by whether their caption names the term, in AUC. CAPTOR improves steadily with $K$
and leads ProxyCLIP from $K=640$ upward. Because every image is scored, these values
are comparable with each other but not with the held-out results of
Table~\ref{tab:cocoseg}.}
\label{tab:coco-k}
\begin{fitwidth}
\begin{tabular}{lccccc}
\toprule
$K$ & $320$ & $640$ & $1280$ & $1920$ & $2560$ \\
\midrule
mIoU & $0.490$ & $0.532$ & $0.541$ & $0.556$ & $0.561$ \\
\makecell[l]{CAPTOR $-$\\ ProxyCLIP} & $-0.016$ & $+0.025$ & $+0.035$ & $+0.050$ & $+0.055$ \\
Label-free score & $0.9539$ & $0.9570$ & $0.9592$ & $0.9598$ & $0.9607$ \\
\bottomrule
\end{tabular}
\end{fitwidth}
\end{table}

\begin{table}[!htp]\centering
\caption{How much each step of the method adds on COCO, scored on the $2{,}500$
reporting images. Each row adds one change to the row above, and the last row is the
configuration reported throughout the paper. \emph{Joint words}, \emph{trust
weights} and \emph{$16$ words per patch} are Equations~\eqref{eq:joint},
\eqref{eq:trust} and~\eqref{eq:softscore}, and \emph{reweighting} is
Equation~\eqref{eq:em}. These steps of the method are worth $+0.032$ between them.
The last two rows change only how captions are read (Section~\ref{sec:s-setup}), and
their gain falls on a few terms. ``Teddy bear'' rises from $0.001$ to $0.706$ mIoU
and ``hot dog'' from $0.419$ to $0.609$, because every caption mentioning a bear or a
dog had counted as evidence for them. Requiring the full name of every multi-word
term instead is worse, at $0.543$, since captions write ``table'' and ``plant''
rather than ``dining table'' and ``potted plant''.}
\label{tab:coco-steps}
\begin{fitwidth}
\begin{tabular}{lcc}
\toprule
Configuration & mIoU & Acc. \\
\midrule
Four vocabularies averaged, one word per patch & $0.503$ & $0.714$ \\
$+$ joint words, trust weights, $16$ words per patch & $0.525$ & $0.739$ \\
$+$ one round of reweighting & $0.535$ & $0.755$ \\
$+$ phrase-aware caption reading & $0.554$ & $0.767$ \\
$+$ full name required for eight terms & $\mathbf{0.565}$ & $\mathbf{0.768}$ \\
\bottomrule
\end{tabular}
\end{fitwidth}
\end{table}

\begin{table}[!htp]\centering
\caption{Alternative ways to turn the counts into a term's signature, in mIoU at $K=1280$ on all
$5{,}000$ COCO validation images. Each row replaces the difference of means of
Equation~\eqref{eq:signature} and keeps the same visual words, propagation and
read-out. \emph{Point-biserial correlation} correlates a word's share of each image with
whether the caption names the term. \emph{Share of the word in images naming the
term} is its mean share in those images, and \emph{lift} divides that by the word's
mean share over the whole corpus. The statistic of Mori et al.~\cite{mori1999} is the
fraction of the word's patches that lie in images naming the term. The \emph{ridge
fit} regresses mentions on word shares in closed form, at the normalization it does
best with, and the \emph{joint estimate} is Equation~\eqref{eq:jointest}. \emph{Raw}
uses each statistic as computed, and \emph{divided by its null} divides it by its own
permutation null, drawn with the same shuffles. \emph{Welch's $t$} divides the
difference of means by its analytic standard error. Dividing by the null raises every statistic except the ridge fit and Welch's $t$, which is already scaled, and Mori's share is best here, so Table~\ref{tab:morireported}
tests it in the full pipeline.}
\label{tab:estimators}
\begin{fitwidth}
\begin{tabular}{lcc}
\toprule
Statistic & Raw & Divided by its null \\
\midrule
Difference of means, Eq.~\eqref{eq:signature} & $0.452$ & $0.541$ \\
Point-biserial correlation & $0.508$ & $0.541$ \\
Welch's $t$, analytic standard error & $0.526$ & $0.364$ \\
Lift of the word under the term & $0.449$ & $0.510$ \\
Share of the word in images naming the term & $0.449$ & $0.493$ \\
Share of the word's patches naming the term~\cite{mori1999} & $0.505$ & $\mathbf{0.569}$ \\
Ridge fit of mentions on word shares & $0.541$ & $0.456$ \\
Joint estimate, ridge $0.01$ & $0.477$ & $0.538$ \\
\bottomrule
\end{tabular}
\end{fitwidth}
\end{table}

\begin{table}[!htp]\centering
\caption{Whether the statistic of Mori et al.~\cite{mori1999}, the best in
Table~\ref{tab:estimators}, also wins in the full reported COCO pipeline of four
vocabularies of $640$ words with joint words, trust weights, reweighting and
propagation. \emph{Difference of means} is Equation~\eqref{eq:signature}, and
$P(t \mid k)$ is the fraction of word $k$'s patches in images whose caption names
term $t$. Each runs over the same grid of settings, $\tau \in \{0.3, 1, 3\}$,
$\lambda \in \{1, 3, 10, 30\}$, $m \in \{1, 16\}$ and $\beta \in \{3, 10\}$, chosen
on the development half, and the reporting half is read once. The last two rows are
Mori minus the difference of means, with intervals that resample reporting images.
The difference of means is higher in mIoU and Mori's in accuracy, so the paper keeps
the difference of means. On this grid the signature's chosen setting, $\tau=3$,
$\lambda=10$ and $m=16$, gives $0.567$, against the $0.565$ of the setting reported in
the paper, and Mori's is $\tau=3$, $\lambda=30$ and $m=1$.}
\label{tab:morireported}
\begin{fitwidth}
\begin{tabular}{lccc}
\toprule
Estimator & \makecell{Development\\mIoU} & \makecell{Reporting\\mIoU} & \makecell{Reporting\\accuracy} \\
\midrule
Difference of means & $\mathbf{0.576}$ & $\mathbf{0.567}$ & $0.771$ \\
Mori et al.~\cite{mori1999}, $P(t \mid k)$ & $0.563$ & $0.555$ & $\mathbf{0.806}$ \\
\midrule
\makecell[l]{Mori $-$\\ difference of means} & & $-0.012$ & $+0.035$ \\
 & & $[-0.019, -0.003]$ & $[+0.028, +0.043]$ \\
\bottomrule
\end{tabular}
\end{fitwidth}
\end{table}

\begin{table}[!htp]\centering
\caption{Whether estimating all terms jointly (Equation~\eqref{eq:jointest}) separates terms
that appear together, COCO at $K=1280$. A microwave is almost always photographed
beside an oven, so when each term is counted separately, the oven's visual words also
become evidence for ``microwave''. Counting all terms at once lets each companion
claim its own words. \emph{Separate} is the
difference of means, and the \emph{joint} rows use the ridge $\lambda_J$ given. Each pair column follows one class on the reporting half and gives the share of its
cells labeled correctly. In brackets is the share labeled as its most frequent
companion, such as a microwave labeled ``oven''. The joint estimate moves every pair toward the right label, most for microwave, whose
correct share rises from $0.25$ to $0.37$ while its share labeled ``oven'' falls from
$0.64$ to $0.49$. It slightly lowers mIoU on the development half, however, so COCO
keeps the separate estimate. On VOC the development half
selects it (Table~\ref{tab:voc}).}
\label{tab:joint}
\begin{fitwidth}
\setlength{\tabcolsep}{3pt}
\begin{tabular}{lccccc}
\toprule
Estimate & \makecell{mIoU,\\development} & \makecell{mIoU,\\reporting} & \makecell{Microwave\\(as oven)} & \makecell{Potted plant\\(as vase)} & \makecell{Cup\\(as dining table)} \\
\midrule
Separate & $\mathbf{0.550}$ & $\mathbf{0.532}$ & $0.25$ $(0.64)$ & $0.23$ $(0.50)$ & $0.21$ $(0.62)$ \\
Joint, $0.01$ & $0.548$ & $0.526$ & $0.34$ $(0.54)$ & $0.25$ $(0.51)$ & $0.24$ $(0.56)$ \\
Joint, $0.001$ & $0.542$ & $0.522$ & $0.37$ $(0.49)$ & $0.26$ $(0.52)$ & $0.25$ $(0.50)$ \\
\bottomrule
\end{tabular}
\end{fitwidth}
\end{table}

\begin{table}[!htp]\centering
\caption{What gives the trained alignment Talk2DINO its edge on COCO, the larger
input or the pixel-adaptive refinement its released configuration applies. All rows
use the same reporting images and each method's released code, and \emph{input} is
the image side in pixels. The refinement is generic post-processing taken from
weakly supervised segmentation, and it adds the cost shown in Table~\ref{tab:cost}.
Applying the same refinement to CAPTOR's map, read here in single precision at
$0.566$, gives $0.570$ mIoU, a paired difference of $+0.004$ with an interval from
$-0.003$ to $+0.009$ that contains zero. The released configuration leads the unrefined CAPTOR by $0.022$, with an interval from $+0.010$ to $+0.037$. About $0.014$ of the gap comes from the refinement, which raises Talk2DINO by $0.013$ at both input sizes but CAPTOR by only $0.004$, and about $0.009$ from the larger input. At the input CAPTOR reads and without refinement, the two are level.}
\label{tab:t2d}
\begin{fitwidth}
\begin{tabular}{lcccc}
\toprule
 & Input & Refinement & mIoU & Acc. \\
\midrule
CAPTOR & $196$ & No & $\mathbf{0.565}$ & $\mathbf{0.768}$ \\
Talk2DINO & $196$ & No & $0.566$ & $0.746$ \\
Talk2DINO & $196$ & Yes & $0.579$ & $0.759$ \\
Talk2DINO & $448$ & No & $0.575$ & $0.753$ \\
Talk2DINO & $448$ & Yes & $\mathbf{0.588}$ & $0.764$ \\
\bottomrule
\end{tabular}
\end{fitwidth}
\end{table}

\FloatBarrier

\section{Ranking whole images for a term}
\label{sec:s-rank}

\begin{figure}[!htp]\centering
\includegraphics[width=0.85\linewidth]{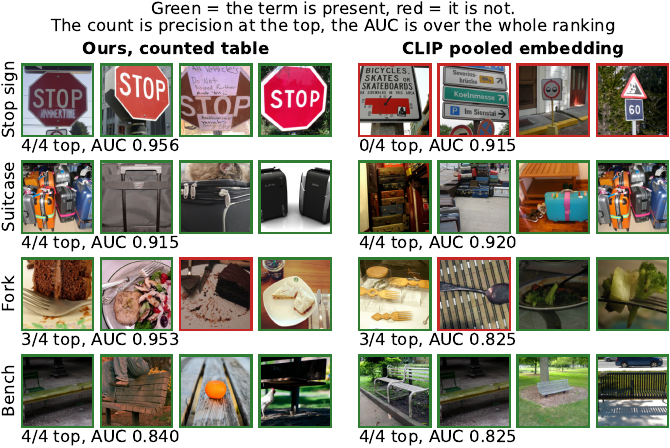}
\caption{Which images each method ranks highest for a term. Each row is one term. The left four images are CAPTOR's top four, and the right four are the top four of
the released CLIP model. A green border means the image contains the term, and a red
one means it does not.
Terms are spread evenly from rare to common.}
\label{fig:cocoattn}
\end{figure}

\begin{figure*}[!htp]\centering
\includegraphics[width=0.49\linewidth]{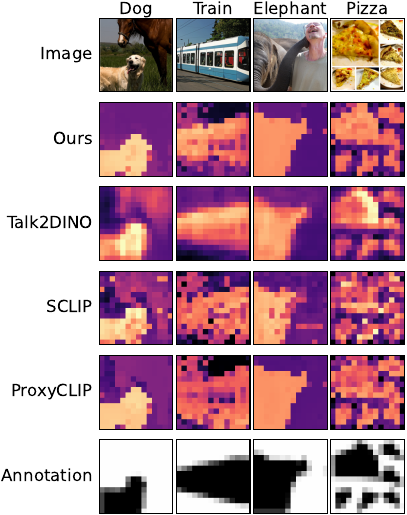}
\caption{Score maps for one term at a time. Each row shows an image containing the term,
CAPTOR's map, the same term read by Talk2DINO, SCLIP and ProxyCLIP, and the
annotation. Each image is the first validation image in which the term covers between
a fifth and two thirds of the center crop, a rule fixed before drawing. Every map is
rescaled over its own patches, so the colors compare shapes rather than
strengths.}
\label{fig:querymaps}
\end{figure*}

The image-level scoring rule of Equation~\eqref{eq:score} ranks the $5{,}000$ COCO
evaluation images for each term (Table~\ref{tab:cocoattn} and
Figure~\ref{fig:cocoattn}), and Figure~\ref{fig:querymaps} draws one row of the
matrix per query. The same rule ranks pathology fields in
Section~\ref{sec:s-path}.

Figure~\ref{fig:highres} carries no annotation, so it illustrates rather than
measures. Read as one term at each cell, six of the ten photographs take their own
term as the most common object term. The other four lose it to something the
photograph really contains, the person holding the dog, the table under the pizza,
snow read as ``skis'' and ``truck'' for the rusted car.

\begin{table}[!htp]\centering
\caption{Whether the table of signatures can also find which images contain a term. For every
term, all $5{,}000$ COCO evaluation images are ranked, and AUC measures how often an
image containing the term ranks above one that does not. CAPTOR scores an image by
how much of it falls in each visual word, weighted by the term's signature
(Equation~\eqref{eq:score}). The baseline is how CLIP itself ranks images, comparing
its single image embedding with the term's text embedding averaged over eighty prompt
templates. \emph{Stored per image} is what each method keeps for an image, and
\emph{ahead on} counts the classes where CAPTOR ranks better. CAPTOR reaches $0.947$
mean AUC against $0.903$, ahead on $54$ of $77$ classes, a paired difference of
$+0.044$ with interval from $+0.040$ to $+0.048$. At $K=640$ the difference is
$+0.043$, so it does not depend on the vocabulary size. So word counts rank images
better than CLIP's own embedding.}
\label{tab:cocoattn}
\begin{fitwidth}
\begin{tabular}{lccc}
\toprule
Method & Stored per image & Mean AUC & Ahead on \\
\midrule
CAPTOR & $K$ word counts & $\mathbf{0.947}$ & $54$ of $77$ \\
CLIP image embedding & $512$ decimal numbers & $0.903$ & \\
\bottomrule
\end{tabular}
\end{fitwidth}
\end{table}

\FloatBarrier

\section{Transfer to Pascal VOC}
\label{sec:s-voc}
\label{sec:voc}

Pascal VOC 2012~\cite{voc} tests whether a table counted on COCO captions transfers
to other photographs. Nothing is re-estimated, and VOC supplies only images and
annotation. Nineteen of its twenty categories bind on the COCO captions. Every
method is scored on the same $82{,}058$ annotated cells, and the $1{,}449$
validation images are split in half, one half to choose the configuration and the
other to report. No prompts are tuned for VOC, and CAPTOR reads one backbone where
ProxyCLIP reads two.

\begin{table}[!htp]\centering
\caption{Which CAPTOR setting works best when the table counted on COCO captions is used, unchanged, on Pascal VOC. The test uses the $1{,}449$ VOC validation images and the
$19$ of $20$ categories the COCO captions cover. Each row is one setting, a vocabulary
size $K$ and a number of COCO captions counted. \emph{Development} and
\emph{reporting} are CAPTOR's mIoU on the two halves of the images, \emph{ProxyCLIP}
is ProxyCLIP's reporting mIoU with the same propagation, and the last column is the
difference. The development half picks the starred row. Among the plain settings it
would pick $K=640$ counted on $30{,}000$ captions, and every interval against
ProxyCLIP, resampling images, contains zero, the widest running from $-0.050$ to
$+0.014$. Adding the joint estimate of Section~\ref{sec:m-refine} to that setting,
with every COCO term as an extra term, raises the development half from $0.827$ to
$0.847$, so it is picked. It leads ProxyCLIP with an interval from $+0.005$ to
$+0.051$, and SCLIP, at $0.750$, by $+0.079$ with an interval from $+0.056$ to
$+0.104$. The label-free rule of Section~\ref{sec:setup}, which reads only the
captions, would pick $K=960$, the weakest of the settings counted on $30{,}000$ captions, so that rule does not carry
over to a new dataset.}
\label{tab:voc}
\begin{fitwidth}
\begin{tabular}{llcccc}
\toprule
$K$ & \makecell{Captions\\counted} & Development & Reporting & ProxyCLIP & \makecell{CAPTOR $-$\\ProxyCLIP} \\
\midrule
$1280$ & $15{,}000$ & $0.819$ & $0.788$ & $0.801$ & $-0.013$ \\
$1920$ & $15{,}000$ & $0.810$ & $0.776$ & $0.801$ & $-0.026$ \\
$320$ & $30{,}000$ & $0.825$ & $0.813$ & $0.801$ & $+0.011$ \\
$640$ & $30{,}000$ & $0.827$ & $0.807$ & $0.801$ & $+0.005$ \\
$960$ & $30{,}000$ & $0.810$ & $0.783$ & $0.801$ & $-0.018$ \\
\midrule
$640$, joint$^{\star}$ & $30{,}000$ & $0.847$ & $\mathbf{0.829}$ & $0.801$ & $+0.027$ \\
\bottomrule
\end{tabular}
\end{fitwidth}
\end{table}

\begin{table}[!htp]\centering
\caption{How CAPTOR compares on the $725$ VOC reporting images with Talk2DINO, a trained
method that learns to align text and image features from COCO captions. CAPTOR uses
the setting Table~\ref{tab:voc} picks, and Talk2DINO uses its own prompts for each
class, on the same images and cells. \emph{Our input} is $196$ pixels,
\emph{propagated} adds CAPTOR's propagation, and \emph{as released} is $448$ pixels
with a pixel-adaptive refinement. The last column is CAPTOR minus the row, with $95\%$
intervals that resample images. CAPTOR is level with Talk2DINO at our input size, with
or without CAPTOR's propagation, and the released configuration leads by $0.022$ with
an interval that only just excludes zero.}
\label{tab:voct2d}
\begin{fitwidth}
\begin{tabular}{lcc}
\toprule
Method & mIoU & CAPTOR $-$ row \\
\midrule
CAPTOR & $0.829$ & \\
Talk2DINO, our input & $0.839$ & \makecell{$-0.011$\\ $[-0.030, +0.008]$} \\
Talk2DINO, our input, propagated & $0.844$ & \makecell{$-0.016$\\ $[-0.035, +0.003]$} \\
Talk2DINO, as released & $0.850$ & \makecell{$-0.022$\\ $[-0.042, -0.0002]$} \\
\bottomrule
\end{tabular}
\end{fitwidth}
\end{table}

\begin{figure*}[!htp]\centering
\includegraphics[width=\linewidth]{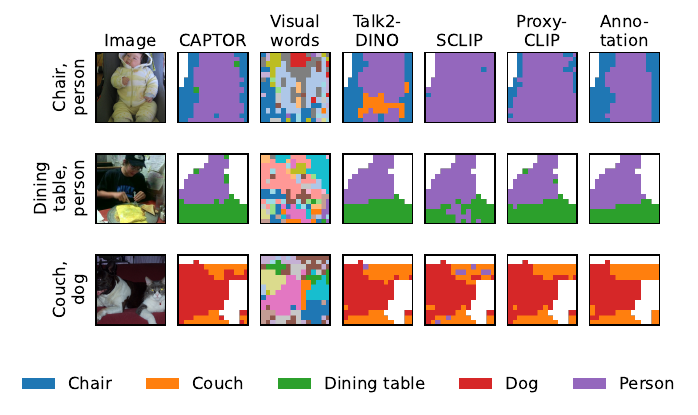}
\caption{Pascal VOC read with the table counted on COCO captions, with nothing re-estimated
for VOC and the joint estimate the development half picks. The three images are those
on which the methods disagree most, among validation images with two or more terms
each covering over a quarter of the center crop. This rule never looks at which method
is right, so it can show rows that favor any method. Columns are the image, the term
CAPTOR scores highest, the visual words themselves without names, Talk2DINO, SCLIP and
ProxyCLIP from their released code, and the annotation. Only the cells the annotation
labels, which are the ones scored, are colored. In the first row SCLIP gives almost
the whole image to ``person'' and misses the chair that CAPTOR and ProxyCLIP both
find.}
\label{fig:voc}
\end{figure*}

\FloatBarrier

\section{COCO-Object under the standard protocol}
\label{sec:s-cobj}
\label{sec:supp-cobj}

\begin{figure*}[!htp]\centering
\includegraphics[width=\linewidth]{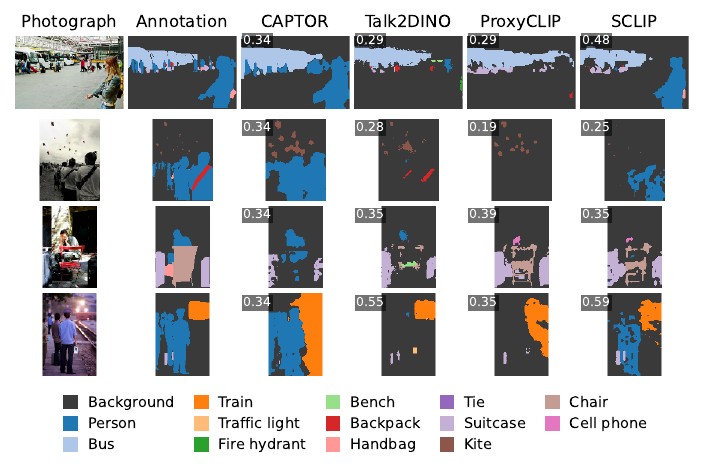}
\caption{COCO-Object images from the reporting half, chosen as typical, with every pixel
scored and background as an eighty-first class. The four images were picked before
any prediction was drawn, as those with two or more object classes whose CAPTOR mIoU
is closest to the median of that pool, $0.344$. Every method runs with the code and
setting of its row in Table~\ref{tab:cobj}, and each number is that image's mIoU.
CAPTOR is ahead on the second image and behind at least one baseline on the other
three. Over the four images it gives the right name to $70\%$ of object pixels,
against $26\%$ to $53\%$ for the baselines, but calls $20\%$ of the background an
object, against $3\%$ to $8\%$. In the last row ``train'' spreads over the platform,
the error Section~\ref{sec:s-cobj} measures.}
\label{fig:cobjqual}
\end{figure*}

COCO-Object scores every pixel of the $5{,}000$ COCO validation images, with
background as an eighty-first class (Table~\ref{tab:cobj} and
Figure~\ref{fig:cobjqual}). Every baseline
reproduces its published number within $0.005$. CAPTOR reaches $0.350$ on the
reporting half. That is above SCLIP at $0.335$ and $0.023$ below ProxyCLIP as
released, at $0.373$, and Talk2DINO as released leads, at $0.448$. Given the same
propagation as CAPTOR, ProxyCLIP falls to $0.353$. The gap lies in where objects
end rather than in what they are. Of all annotated object pixels, CAPTOR gives the
right name to $0.632$, against $0.462$ for SCLIP, $0.429$ for ProxyCLIP and $0.450$
for Talk2DINO. The baselines call about half of all object pixels background.
CAPTOR does the opposite and lets objects spread into the background around them,
which mIoU penalizes. Its background reaches an IoU of $0.676$, against $0.782$ for
Talk2DINO. COCO-Object's background is everything outside eighty object classes,
and scene names are the terms a caption corpus supplies least well.

To recognize background, SCLIP adds a list of $26$ scene names, such as ``sky'',
``road'' and ``wall'', and a pixel that best matches one of them is called background.
CAPTOR is given the same list except ``bed'', which is also one of the eighty object
classes and so cannot stand for background as well. Its background threshold of
$2.5$ is chosen on the development half.
Without scene names CAPTOR reaches only $0.261$, and read at the $196$ pixels its
words were counted from it reaches $0.292$. ``Toaster'' and ``hair drier'' are too
rare to count and score zero.

On the first $500$ images, CAPTOR calls $13.2\%$ of background pixels an object
more than $56$ pixels from any object, against $2.0\%$ for Talk2DINO. The names it
paints there belong to objects with a typical setting, such as ``surfboard'',
``skis'' and ``sink'', whose visual words absorb that setting. SCLIP's scene list
has no ``snow'', ``water'' or ``tile'' to claim them.

Two corrections chosen on the development half narrow this gap. Scene names from
the $91$ COCO-Stuff classes raise the reporting half from $0.350$ to $0.356$, and
weighting the affinity of Equation~\eqref{eq:affinity} by the distance between
cells adds $0.002$, for $0.358$ together. Estimating objects and scene terms
jointly weakens the objects, at $0.333$ on the development half. The baselines keep
the paper's propagation, so this is an analysis of CAPTOR and not a comparison.

\begin{table*}[!htp]\centering
\caption{Whether each baseline's published COCO-Object score can be reproduced, and where CAPTOR stands. COCO-Object scores every pixel of the $5{,}000$ validation images, with
background as an eighty-first class. \emph{Published} is the score in each method's
paper, or for SCLIP in the repository whose code we run, since its paper reports
$0.305$ for an earlier version. \emph{Reproduced} is our run of the released code on
all images. The last two columns score only the reporting half, first as released and
then with CAPTOR's whole-image propagation. CAPTOR is read at $196$ pixels or at each
image's full size, and its last row adds the background threshold chosen on the
development half. \emph{Caption pairs} is how much image--caption data each text side
was learned from, the $400$M pairs of CLIP, plus $83$k COCO captioned images for the
projection Talk2DINO trains, against the $15$k captions CAPTOR counts. Bold marks the
best value in each column. Every baseline reproduces its published number within $0.005$.
CAPTOR reaches $0.350$, above SCLIP and below ProxyCLIP and Talk2DINO, and
Section~\ref{sec:s-cobj} shows that the gap comes from background pixels.}
\label{tab:cobj}
\begin{fitwidth}
\begin{tabularx}{\linewidth}{>{\raggedright\arraybackslash}Xccccc}
\toprule
Method & \makecell{Caption\\pairs} & \makecell{Pub-\\lished} & \makecell{Repro-\\duced} & \makecell{Reporting\\half} & \makecell{Reporting\\half,\\propagated} \\
\midrule
\multicolumn{6}{l}{\emph{COCO-Object, $5{,}000$ images}} \\
SCLIP~\cite{sclip} & $400$M & $0.335$ & $0.337$ & $0.335$ & $0.346$ \\
ProxyCLIP~\cite{proxyclip} & $400$M & $0.375$ & $0.371$ & $0.373$ & $0.353$ \\
Talk2DINO, no refinement~\cite{talk2dino} & \makecell{$400$M\\$+\,83$k} & $0.410$ & $0.412$ & $0.408$ & $\mathbf{0.439}$ \\
Talk2DINO, as released~\cite{talk2dino} & \makecell{$400$M\\$+\,83$k} & $\mathbf{0.451}$ & $\mathbf{0.451}$ & $\mathbf{0.448}$ & $0.434$ \\
CAPTOR, at $196$ pixels & $15$k & & & $0.292$ & \\
CAPTOR, at native resolution & $15$k & & & $0.318$ & $0.317$ \\
CAPTOR, native, development threshold & $15$k & & & & $0.350$ \\
\bottomrule
\end{tabularx}
\end{fitwidth}
\end{table*}

\FloatBarrier

\section{Pathology}
\label{sec:s-path}
\label{sec:supp-path}

\begin{figure}[!htp]\centering
\includegraphics[width=\linewidth]{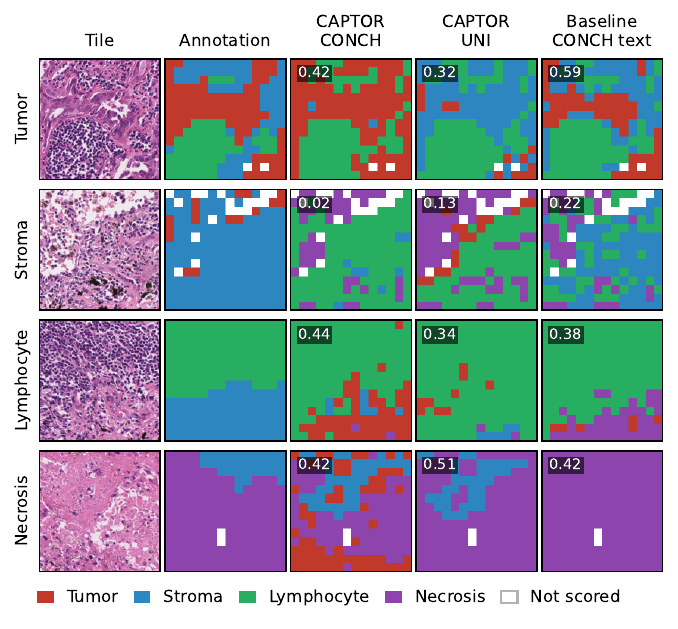}
\caption{Lung tiles from the LUAD-HistoSeg reporting patients, chosen as typical, with the
settings of Table~\ref{tab:enctissue}. For each tissue, the row shows the tile, among
those mostly of that tissue, on which CAPTOR on CONCH scores the median. Maps are
drawn on the $14\times14$ grid that is scored, and each number is that tile's mIoU.
White cells are less than half annotated and are not scored. The baseline reads
CONCH's text encoder with CAPTOR's propagation and standardization. CAPTOR on CONCH
leads on the lymphocyte tile, is level on necrosis and falls behind on tumor and
stroma. On the stroma tile both CAPTOR rows name most cells ``lymphocyte'' or
``necrosis'', while the baseline names about half of them stroma.}
\label{fig:luadqual}
\end{figure}

This section gives every BCSS baseline, the result per tissue and per component,
the configurations behind Table~\ref{tab:encoders} with typical lung tiles
(Figure~\ref{fig:luadqual}), stain robustness, localization, field ranking and
naming in further organs and encoders.

We tried a range of trust weights $\lambda$ and reweighting settings on the
development patients. The best choice was a very large $\lambda$ with no reweighting,
which reduces to the averaged scoring rule, and it reaches $0.469$ mIoU on held-out
patients. Forcing the natural-image setting lowers this to
$0.455$. That experiment drew its own random set of fields, covering $98$ of the $103$
patients, so its $0.469$ cannot be compared directly with the $0.497$ of
Table~\ref{tab:pathmain}. Reweighting has little to act on, since only
$5.7\%$ of the captions name two or more of the five terms.

Table~\ref{tab:naming6} tests how well the $K=40$ visual words of six pathology
encoders are named. Four were trained with captions, CONCH~\cite{conch},
PLIP~\cite{plip}, QuiltNet~\cite{quilt} and BiomedCLIP~\cite{biomedclip}, and two without, UNI~\cite{uni} and Phikon~\cite{phikon}. UNI and Phikon have no
text encoder of their own, so their visual words can be named only by counting, as
CAPTOR does, or by borrowing another model's text encoder. CAPTOR names a word by the term
whose signature is largest on it, using only that encoder and the captions. The two
text-encoder namers embed the $60$ PathCap figures where the word is most common
with CONCH and choose the closest of twelve tissue terms, using CONCH's text
encoder or the adapted CLIP encoder. Each name is scored by AUC on labelled tiles
from NCT-CRC~\cite{nct}, LUAD-HistoSeg~\cite{luadhistoseg} and
PanNuke~\cite{pannuke} that were never used to choose a name.

Against the adapted CLIP encoder, CAPTOR is ahead in $12$ of the $18$ cells, and the mean difference is $-0.002$, so the two are level. CAPTOR leads in $10$ of $12$ cells on the encoders trained with captions and
in $2$ of $6$ on UNI and Phikon. Against CONCH's released text encoder, it leads in
all nine cells of CONCH, QuiltNet and BiomedCLIP and trails in all nine of PLIP,
UNI and Phikon. Both text-encoder namers see the figures through CONCH, which
CAPTOR on UNI or Phikon does not use.

The label-free rule for $K$ ranks held-out captions by their signatures. On
PathCap it gives AUC $0.7830$, $0.7964$ and $0.8095$ at $K=40$, $80$ and $320$, so
it selects $K=320$, which the development patients rank worst. Separating captions
keeps improving as $K$ grows, while segmenting five tissues peaks earlier.

\begin{table}[!htp]\centering
\caption{How CAPTOR compares with every other method that labels tissue patch by patch
using the same image encoder, CONCH. All $200$ BCSS fields from $103$ patients are scored on a $28\times28$
grid with five tissue terms, at $K=40$ and without propagation, so every method is
compared in its plain form. A dagger marks methods that use CONCH's text encoder.
\emph{Dense} (MaskCLIP) compares each patch with the term's text embedding, and
\emph{region-pooled} first pools the patches of each grid cell. \emph{Self-affinity}
averages dense scores over similar-looking patches, as CLIP-DINOiser does, and
\emph{re-encoding} runs the encoder again on sixty-four crops per field. \emph{On our
words} averages a method's scores within each of CAPTOR's visual words, and \emph{soft
assignment} lets a patch belong to several words (Equation~\eqref{eq:softscore}).
\emph{Standardized} puts each term's scores on a common scale without reading any
label. CAPTOR averages four vocabularies and leads the strongest baseline that does
not borrow its words by $+0.060$. The two hybrids on its words lead CAPTOR by $0.030$
to $0.035$, so the visual words carry the gain. Each grid cell takes its most common annotated class, and cells whose class is not one of the five tissues are not scored. A single vocabulary ranges from $0.260$ to $0.394$ on these fields when each cell instead takes its most common of the five tissues, as in Table~\ref{tab:pathabl}, which is why four are averaged. \emph{Most common
tissue everywhere} always guesses the most frequent tissue, and the \emph{supervised
probe}, trained on the labels, is an upper reference, not a method.}
\label{tab:desc}
\begin{fitwidth}
\begin{tabular}{lcc}
\toprule
Method & mIoU & Accuracy \\
\midrule
Region-pooled on our words, standardized$^{\dagger}$ & $0.404$ & $0.668$ \\
Dense on our words, standardized$^{\dagger}$ & $0.399$ & $0.664$ \\
CAPTOR, four vocabularies & $0.369$ & $0.644$ \\
CAPTOR, soft assignment, standardized & $0.351$ & $0.583$ \\
SCLIP, standardized$^{\dagger}$ & $0.309$ & $0.506$ \\
Self-affinity, standardized$^{\dagger}$ & $0.306$ & $0.518$ \\
Region-pooled, standardized$^{\dagger}$ & $0.306$ & $0.519$ \\
Region-pooled$^{\dagger}$ & $0.290$ & $0.487$ \\
Self-affinity$^{\dagger}$ & $0.286$ & $0.479$ \\
Dense MaskCLIP$^{\dagger}$ & $0.264$ & $0.460$ \\
SCLIP$^{\dagger}$ & $0.262$ & $0.436$ \\
Re-encoding, standardized$^{\dagger}$ & $0.168$ & $0.344$ \\
Most common tissue everywhere & $0.085$ & $0.426$ \\
Re-encoding$^{\dagger}$ & $0.060$ & $0.093$ \\
\midrule
Supervised probe & $0.612$ & $0.836$ \\
\bottomrule
\end{tabular}
\end{fitwidth}
\end{table}

\begin{table}[!htp]\centering
\caption{Table~\ref{tab:desc} broken down by tissue, in IoU on the same fields. The first
three rows are CAPTOR with one vocabulary at $K=40$ and two baselines from that table,
and bold marks the best of the three. CAPTOR is best on four of the five tissues, and region-pooled scoring leads on necrosis, $0.283$ against $0.266$. The
fourth row is the configuration reported in Table~\ref{tab:pathmain}, $K=80$ with four
vocabularies and the propagation, at $0.494$ mIoU and $0.735$ accuracy on all $200$ fields, with the cell labels of Table~\ref{tab:desc}. Its largest gain over one vocabulary is on necrosis. The supervised probe,
trained on the labels, is an upper reference.}
\label{tab:desc2}
\setlength{\tabcolsep}{2.7pt}
\begin{tabular}{lccccc}
\toprule
& Tumor & Stroma & Lymphocyte & Necrosis & Adipose \\
\midrule
CAPTOR, one vocabulary & $\mathbf{0.670}$ & $\mathbf{0.497}$ & $\mathbf{0.434}$ & $0.266$ & $\mathbf{0.125}$ \\
Region-pooled, standardized & $0.420$ & $0.395$ & $0.377$ & $\mathbf{0.283}$ & $0.056$ \\
Dense (MaskCLIP) & $0.267$ & $0.456$ & $0.331$ & $0.146$ & $0.122$ \\
\midrule
CAPTOR, reported configuration & $0.738$ & $0.470$ & $0.502$ & $0.577$ & $0.181$ \\
\midrule
Supervised probe & $0.810$ & $0.683$ & $0.584$ & $0.679$ & $0.304$ \\
\bottomrule
\end{tabular}
\end{table}

\begin{table}[!htp]\centering
\caption{What each part of the pathology method adds on BCSS, in mIoU on a $28\times28$ grid. Here each cell takes its most common of the five tissues, ignoring other classes, which is why the configuration that reads $0.494$ in Table~\ref{tab:desc2} reads $0.486$ here. \emph{Single} is the average score of four vocabularies, each used on its own,
and \emph{ensemble} combines the four into one score map. The first three rows use all
$200$ fields from $103$ patients, and the last row is the held-out result of
Table~\ref{tab:pathmain} on the $62$ reporting patients. Combining the vocabularies
adds $0.02$ to $0.06$, and at $K=80$ the propagation adds $0.07$ to the ensemble and
$0.11$ to one vocabulary. The propagation helps CAPTOR and the domain text encoder
about equally, $+0.051$ and $+0.052$ at $K=40$, so it is not what separates them. On
the $66$ development fields from $41$ patients, the combined vocabularies reach
$0.429$, $0.452$ and $0.429$ mIoU at $K=40$, $80$ and $320$. A middle $K$ is best here
because five tissue terms need far fewer words than the seventy-eight terms of
COCO.}
\label{tab:pathabl}
\begin{fitwidth}
\begin{tabular}{llccc}
\toprule
Patients & $K$ & Propagation & Single & Ensemble \\
\midrule
All $103$ & $40$ & Yes & $0.427$ & $0.451$ \\
All $103$ & $80$ & No  & $0.355$ & $0.414$ \\
All $103$ & $80$ & Yes & $0.467$ & $0.486$ \\
Held-out $62$ & $80$ & Yes & $0.475$ & $\mathbf{0.497}$ \\
\bottomrule
\end{tabular}
\end{fitwidth}
\end{table}

\begin{table}[!htp]\centering
\caption{Table~\ref{tab:encoders} broken down by tissue, to show where each encoder gains
or loses. Entries are IoU per tissue, mIoU and pixel accuracy (\emph{acc.}) on the reporting
patients, $62$ of $103$ for BCSS and $15$ of $25$ for LUAD-HistoSeg. All use the
$14\times14$ grid the three encoders share. LUAD-HistoSeg has no adipose class. Each
method picks its settings on development patients. CAPTOR on CONCH uses $K=80$ on
breast and $K=40$ with the joint estimate on lung. CAPTOR on UNI and on Phikon use
$K=80$ with the joint estimate and standardization, and the baseline is dense scoring
with propagation and standardization. On lung, CAPTOR on CONCH labels most
tumor-associated stroma as ``tumor'', at $0.125$ stroma IoU against $0.444$ for the
baseline, while CAPTOR on UNI, counted from the same captions, reaches $0.423$. On the
other three lung tissues CAPTOR on CONCH matches or beats the baseline, so its lung
deficit comes from stroma alone.}
\label{tab:enctissue}
\begin{fitwidth}
\setlength{\tabcolsep}{2.5pt}
\begin{tabular}{lccccccc}
\toprule
Method & Tumor & Stroma & Lymphocyte & Necrosis & Adipose & mIoU & Acc. \\
\midrule
\multicolumn{8}{l}{\emph{breast, BCSS}} \\
CAPTOR on CONCH & $0.766$ & $0.455$ & $0.494$ & $0.644$ & $0.248$ & $\mathbf{0.521}$ & $\mathbf{0.745}$ \\
CAPTOR on UNI & $0.704$ & $0.541$ & $0.507$ & $0.361$ & $0.166$ & $0.456$ & $0.713$ \\
CAPTOR on Phikon & $0.522$ & $0.417$ & $0.410$ & $0.092$ & $0.054$ & $0.299$ & $0.552$ \\
Baseline, CONCH text & $0.463$ & $0.389$ & $0.396$ & $0.298$ & $0.075$ & $0.324$ & $0.539$ \\
\midrule
\multicolumn{8}{l}{\emph{lung, LUAD-HistoSeg}} \\
CAPTOR on CONCH & $0.538$ & $0.125$ & $0.420$ & $0.217$ & & $0.325$ & $0.547$ \\
CAPTOR on UNI & $0.343$ & $0.423$ & $0.574$ & $0.299$ & & $\mathbf{0.410}$ & $0.570$ \\
CAPTOR on Phikon & $0.666$ & $0.393$ & $0.311$ & $0.126$ & & $0.374$ & $\mathbf{0.649}$ \\
Baseline, CONCH text & $0.524$ & $0.444$ & $0.357$ & $0.212$ & & $0.384$ & $0.606$ \\
\bottomrule
\end{tabular}
\end{fitwidth}
\end{table}

\begin{table}[!htp]\centering
\caption{Two separate checks in pathology, in mIoU on the $200$ BCSS fields. \textbf{(a)}
Whether tissue terms can be read without a pathology text encoder, at $K=80$ with
propagation. CAPTOR uses no text encoder, and the other two rows name the tissues
through a general text encoder (CLIP) or through CONCH's pathology text encoder.
CAPTOR scores well above both. Counted from fewer captions and measured without
propagation, it reaches $0.287$ mIoU from $649$ captions, $0.340$ from $1{,}623$ and
$0.396$ from all $32{,}463$, so from $649$ captions it already exceeds CONCH's text encoder measured the same way, without propagation, at $0.276$. \textbf{(b)} Whether stain differences between slides
break the visual words, at $K=40$ with one vocabulary and no propagation. Stain
jitter shifts the colors enough to move $16\%$ of patches to a different visual word,
yet CAPTOR drops only by $0.010$. Dense scoring with CONCH's text encoder is shown for
reference.}
\label{tab:pathextra}
\begin{fitwidth}
\begin{tabular}{lc}
\toprule
Setting & mIoU \\
\midrule
\multicolumn{2}{l}{\emph{(a) reading terms without a pathology text encoder}} \\
General text encoder (CLIP), dense & $0.084$ \\
Domain text encoder (CONCH), standardized & $0.328$ \\
Same, without propagation & $0.276$ \\
CAPTOR, no text encoder & $0.465$ \\
\midrule
\multicolumn{2}{l}{\emph{(b) robustness to stain differences}} \\
CAPTOR, no perturbation & $0.394$ \\
CAPTOR, $16\%$ of patches reassigned & $0.384$ \\
Dense scoring, no perturbation & $0.263$ \\
Dense scoring, $16\%$ of patches reassigned & $0.267$ \\
\bottomrule
\end{tabular}
\end{fitwidth}
\end{table}

\begin{table}[!htp]\centering
\caption{How well each method finds one named tissue at a time. This is easier than
segmentation, because the scores of different terms never need to be compared. Each
tissue is scored patch by patch against the BCSS annotation on the $200$ fields of
Table~\ref{tab:desc}, in AUC, where $0.5$ is chance. \emph{Hard} gives each patch one
visual word and \emph{soft} lets it belong to several (Equations~\eqref{eq:patchscore}
and~\eqref{eq:softscore}), and the other columns are the text-encoder methods of
Table~\ref{tab:desc}. Every method except re-encoding needs one pass of the encoder,
and none uses labels. Intervals over patients run to about $\pm 0.02$, and bold marks
the best method in each row. The soft assignment raises CAPTOR from $0.791$ to $0.867$, level with the best
method, region-pooled scoring, at $0.873$. This
suggests that CAPTOR's lead in Table~\ref{tab:desc} comes from putting all terms on one
scale rather than from finding each tissue better.}
\label{tab:loc2}
\setlength{\tabcolsep}{1.8pt}
\begin{tabular}{lccccccc}
\toprule
Tissue & CAPTOR & CAPTOR & Region- & Self- & SCLIP & Dense & Re- \\
 & Hard & Soft & Pooled & Affinity & & & Encoding \\
\midrule
Adipose & $0.887$ & $0.958$ & $\mathbf{0.985}$ & $0.980$ & $0.980$ & $0.972$ & $0.733$ \\
Necrosis & $0.802$ & $0.875$ & $0.912$ & $0.912$ & $\mathbf{0.940}$ & $0.900$ & $0.794$ \\
Lymphocyte & $0.830$ & $0.909$ & $0.895$ & $0.891$ & $\mathbf{0.920}$ & $0.868$ & $0.630$ \\
Stroma & $0.676$ & $0.753$ & $0.833$ & $\mathbf{0.836}$ & $0.788$ & $0.819$ & $0.603$ \\
Tumor & $0.761$ & $\mathbf{0.838}$ & $0.742$ & $0.724$ & $0.684$ & $0.699$ & $0.677$ \\
\midrule
Mean & $0.791$ & $0.867$ & $\mathbf{0.873}$ & $0.869$ & $0.863$ & $0.851$ & $0.687$ \\
\bottomrule
\end{tabular}
\end{table}

\begin{table}[!htp]\centering
\caption{Whether CAPTOR ranks whole pathology fields by tissue better than a text encoder trained on the same captions. Each field gets one score per term from the image-level rule of
Equation~\eqref{eq:score}, and AUC measures how well the fields containing the tissue
rank above the rest. Each row is one pair of a dataset and a tissue term.
\emph{Released} is CONCH's text encoder as released. \emph{Adapted} is the same encoder
fine-tuned on the caption pairs CAPTOR counts, keeping its weights and the image
encoder frozen and training only rank-$16$ LoRA adapters (Section~\ref{sec:s-setup}). The last
column is CAPTOR minus adapted, with intervals over patients, or over organs for
PanNuke, and bold marks intervals that exclude zero. CAPTOR leads by $+0.087$ on
average, ahead on $12$ of $13$ pairs and clearly ahead on eight, with none clearly
behind. So counting the captions ranks fields better than the text encoder adapted on them here. Against the released encoder CAPTOR leads on seven of the thirteen pairs with a
mean of $+0.038$. Four independently seeded vocabularies give means of $+0.087$,
$+0.083$, $+0.101$ and $+0.086$ over the adapted encoder.}
\label{tab:attnterms}
\begin{fitwidth}
\begin{tabular}{llcccc}
\toprule
Corpus & Term & CAPTOR & Released & Adapted & \makecell{CAPTOR $-$\\ adapted} \\
\midrule
LUAD & Tumor & $0.842$ & $0.678$ & $0.562$ & \makecell{$\mathbf{+0.279}$\\ $[+0.188, +0.364]$} \\
LUAD & Stroma & $0.757$ & $0.608$ & $0.559$ & \makecell{$\mathbf{+0.199}$\\ $[+0.018, +0.387]$} \\
BCSS & Tumor & $0.944$ & $0.643$ & $0.748$ & \makecell{$\mathbf{+0.195}$\\ $[+0.159, +0.231]$} \\
BCSS & Stroma & $0.838$ & $0.824$ & $0.652$ & \makecell{$\mathbf{+0.186}$\\ $[+0.123, +0.254]$} \\
PanNuke & Stroma & $0.666$ & $0.701$ & $0.603$ & \makecell{$\mathbf{+0.063}$\\ $[+0.011, +0.111]$} \\
BCSS & Lymphocyte & $0.908$ & $0.860$ & $0.853$ & \makecell{$\mathbf{+0.055}$\\ $[+0.033, +0.081]$} \\
LUAD & Necrosis & $0.891$ & $0.972$ & $0.838$ & \makecell{$+0.053$\\ $[-0.010, +0.110]$} \\
LUAD & Lymphocyte & $0.968$ & $0.922$ & $0.922$ & \makecell{$\mathbf{+0.047}$\\ $[+0.023, +0.078]$} \\
BCSS & Adipose & $0.968$ & $0.938$ & $0.940$ & \makecell{$\mathbf{+0.028}$\\ $[+0.009, +0.045]$} \\
PanNuke & Tumor & $0.767$ & $0.773$ & $0.751$ & \makecell{$+0.017$\\ $[-0.073, +0.114]$} \\
BCSS & Necrosis & $0.831$ & $0.902$ & $0.825$ & \makecell{$+0.006$\\ $[-0.025, +0.036]$} \\
PanNuke & Epithelium & $0.743$ & $0.790$ & $0.737$ & \makecell{$+0.006$\\ $[-0.037, +0.130]$} \\
PanNuke & Inflammatory & $0.796$ & $0.820$ & $0.796$ & \makecell{$0.000$\\ $[-0.021, +0.041]$} \\
\bottomrule
\end{tabular}
\end{fitwidth}
\end{table}

\begin{table}[!htp]\centering
\caption{Whether the names CAPTOR learns from captions still work in a dataset it never saw. The
test uses the nine expert classes of NCT-CRC colorectal tiles, and everything is fixed
before NCT-CRC is touched. For each term, AUC measures how well it separates the tiles
of its class from the other eight, where $0.5$ is chance. \emph{Name} scores a tile by
how much of it falls in the single visual word named after the term, and
\emph{signature} uses the term's whole signature (Equation~\eqref{eq:score}).
\emph{Best} is the single visual word that separates each class best, chosen by
reading the labels, so it is an upper reference for \emph{name}. \emph{Class code} is
NCT-CRC's own abbreviation, and $\pm$ is the half-width of a $95\%$ interval over
tiles, since this dataset has no patient identifiers. Nine of the twelve names
separate their class with AUC above $0.75$, and the whole signature raises the mean
from $0.787$ to $0.902$, against $0.970$ for the reference.}
\label{tab:nct}
\setlength{\tabcolsep}{2.2pt}
\begin{fitwidth}
\begin{tabular}{llcccc}
\toprule
Term & Class code & Name & $\pm$ & Signature & Best \\
\midrule
Lymphocyte & LYM & $0.999$ & $\pm0.001$ & $0.977$ & $0.999$ \\
Lymphoid & LYM & $0.999$ & $\pm0.001$ & $0.977$ & $0.999$ \\
Adipose & ADI & $0.997$ & $\pm0.004$ & $0.993$ & $0.997$ \\
Necrosis & DEB & $0.874$ & $\pm0.029$ & $0.853$ & $0.907$ \\
Gland & NORM & $0.836$ & $\pm0.025$ & $0.929$ & $0.991$ \\
Glandular & NORM & $0.836$ & $\pm0.025$ & $0.929$ & $0.991$ \\
Carcinoma & TUM & $0.782$ & $\pm0.030$ & $0.941$ & $0.955$ \\
Stroma & STR & $0.764$ & $\pm0.022$ & $0.828$ & $0.931$ \\
Muscle & MUS & $0.753$ & $\pm0.033$ & $0.920$ & $0.932$ \\
\midrule
Tumor & TUM & $0.645$ & $\pm0.033$ & $0.792$ & $0.955$ \\
Epithelium & NORM & $0.504$ & $\pm0.009$ & $0.894$ & $0.991$ \\
Mucin & MUC & $0.451$ & $\pm0.023$ & $0.789$ & $0.993$ \\
\midrule
Mean & & $0.787$ & & $0.902$ & $0.970$ \\
\bottomrule
\end{tabular}
\end{fitwidth}
\end{table}

\begin{table}[!htp]\centering
\caption{Whether counting can name the visual words of other pathology encoders as well as a text encoder can. Each of six encoders has its own $K=40$ visual words, and each method
names every word with one of twelve tissue terms. The names are tested on three
labelled datasets no method saw, colorectal (NCT-CRC), lung (LUAD-HistoSeg) and
19 organs (PanNuke), in mean AUC. CAPTOR names a word from the captions alone.
\emph{CONCH text} and \emph{adapted CLIP} instead look at the PathCap figures where
the word is most common and pick the term whose text embedding is closest. They use
CONCH's released text encoder or a CLIP text encoder trained on the same captions. The
last column is CAPTOR minus adapted CLIP. CAPTOR is ahead in $10$ of $12$ cells on the
four encoders trained with captions and in $2$ of $6$ on UNI and Phikon, which were
trained without text. Overall it is level with the adapted encoder. So counting names any encoder's visual words about as well as a text encoder trained
on the same captions, without gradient training. Of the three routes, it is the only
one that needs nothing beyond the encoder and the captions.}
\label{tab:naming6}
\begin{tabular}{llcccc}
\toprule
Encoder & Dataset & CAPTOR & \makecell{CONCH\\text} & \makecell{Adapted\\CLIP} & \makecell{CAPTOR $-$\\adapted} \\
\midrule
CONCH & Colorectal & $0.835$ & $0.684$ & $0.773$ & $+0.062$ \\
 & Lung & $0.794$ & $0.784$ & $0.790$ & $+0.004$ \\
 & 19 organs & $0.711$ & $0.666$ & $0.720$ & $-0.009$ \\
\midrule
PLIP & Colorectal & $0.714$ & $0.798$ & $0.663$ & $+0.051$ \\
 & Lung & $0.767$ & $0.777$ & $0.756$ & $+0.011$ \\
 & 19 organs & $0.685$ & $0.730$ & $0.657$ & $+0.028$ \\
\midrule
QuiltNet & Colorectal & $0.723$ & $0.700$ & $0.643$ & $+0.080$ \\
 & Lung & $0.658$ & $0.631$ & $0.711$ & $-0.053$ \\
 & 19 organs & $0.640$ & $0.613$ & $0.637$ & $+0.004$ \\
\midrule
BiomedCLIP & Colorectal & $0.823$ & $0.633$ & $0.776$ & $+0.046$ \\
 & Lung & $0.787$ & $0.633$ & $0.729$ & $+0.058$ \\
 & 19 organs & $0.651$ & $0.529$ & $0.628$ & $+0.022$ \\
\midrule
UNI & Colorectal & $0.659$ & $0.686$ & $0.597$ & $+0.062$ \\
 & Lung & $0.648$ & $0.746$ & $0.827$ & $-0.179$ \\
 & 19 organs & $0.666$ & $0.716$ & $0.663$ & $+0.003$ \\
\midrule
Phikon & Colorectal & $0.676$ & $0.764$ & $0.746$ & $-0.070$ \\
 & Lung & $0.691$ & $0.759$ & $0.791$ & $-0.100$ \\
 & 19 organs & $0.555$ & $0.584$ & $0.606$ & $-0.051$ \\
\bottomrule
\end{tabular}
\end{table}

\FloatBarrier

\section{Cost and data efficiency}
\label{sec:s-cost}

Table~\ref{tab:cost} compares the term side of each method on natural images, how a
term is prepared and how much it costs to answer, since the image encoder costs every
method about the same. Table~\ref{tab:cost2} compares building CAPTOR's pathology
table with training the adapted CLIP encoder of Section~\ref{sec:s-setup} on the same
captions. On the same GPU, building CAPTOR's whole table, including $k$-means and the
permutation nulls, takes $4.3$\,s without gradients, against $160.5$\,s and
$6{,}084$ gradient steps to train the adapter. On the CPU alone CAPTOR needs
$57.2$\,s. All times in this section were
measured on the same machine, one NVIDIA RTX 5070 Ti with $16$\,GB, an AMD Ryzen 7
7800X3D, PyTorch $2.10$ and CUDA $12.8$.

\begin{table}[!htp]\centering
\caption{The term side of each method on COCO, measured on the same GPU, since every
method encodes the image with a comparable backbone. \emph{Learned values} are the numbers each method estimates from data, by gradient
descent for the text routes and by clustering and counting for CAPTOR.
\emph{Preparation} is what must happen before the first query. CAPTOR builds its table from the $15{,}000$ captions, after the image encoder pass all
methods share. The text routes use CLIP's text encoder as pretrained on about $400$ million image and
text pairs, and Talk2DINO adds a projection its authors trained on COCO captions. \emph{Inference} makes one new term
queryable, a text encoder pass over $80$ prompts for CLIP and $7$ for Talk2DINO, or counting the captions for CAPTOR. That count covers caption matching from a prebuilt word index, the signatures and their $200$ permutation draws for four vocabularies, and the joint words, but not the caption-conditioned reweighting. So CAPTOR learns its values from captions in seconds, without gradients or a
pretrained text model, and adds a term with about thirty times less computation than
CLIP. Once a term is prepared, scoring an
image against it takes every method a fraction of a millisecond. For scale, encoding
an image takes $2.29$\,ms for CAPTOR and $2.17$\,ms for Talk2DINO at a batch of $32$.
Labelling one image with all $78$ terms takes $8.2$\,ms for CAPTOR against
$387$\,ms for Talk2DINO's released refinement.}
\label{tab:cost}
\setlength{\tabcolsep}{4pt}
\begin{tabular}{lccc}
\toprule
 & CAPTOR & \makecell{CLIP text encoder\\(SCLIP, ProxyCLIP)} & Talk2DINO \\
\midrule
Learned values & $16.1$\,M & $63.4$\,M & $72.0$\,M \\
Preparation & \makecell{$8.6$\,s, counting\\and $k$-means} & \makecell{Pretrained\\by OpenAI} & \makecell{CLIP pretrained,\\projection trained\\on COCO captions} \\
Inference time & $3.2$\,ms & $37.3$\,ms & $6.3$\,ms \\
Inference compute & $15.4$\,GFLOPs & $476.8$\,GFLOPs & $41.7$\,GFLOPs \\
\bottomrule
\end{tabular}
\end{table}

\begin{table}[!htp]\centering
\caption{The time it takes to turn a set of captions into a naming table, the table that gives
each visual word a name, for CAPTOR and for a CLIP text encoder adapted with low-rank adapters. Both first need
the image embeddings of the captioned figures, and that shared step is left out.
Both are timed on the same GPU. The table shows the part that grows with the number
of caption pairs. For CAPTOR this is counting, and for the adapter it is training, with its steps and trained parameters. The adapter trains rank-$16$ LoRA on the query and value projections of CLIP's text encoder and a linear head, with a cosine regression loss, for six epochs or more where needed to reach at least $600$ steps. CAPTOR also runs two steps
only once, building its visual words with $k$-means ($4.2$\,s) and
the permutation nulls ($0.12$\,s). Including them it needs
$4.3$\,s in total, against $160.5$\,s of training and $1.3$\,GB of GPU
memory for the adapter. On the CPU alone CAPTOR needs $57.2$\,s. Training uses batch size $32$, and counting times are the median of three repeats. Counting takes milliseconds, and even with its one-off steps CAPTOR builds the table
far faster than the adapter trains, and still faster on the CPU alone.}
\label{tab:cost2}\label{tab:termscale}
\setlength{\tabcolsep}{4pt}
\begin{tabular}{lcccc}
\toprule
\makecell{Caption\\pairs} & \makecell{CAPTOR\\counting} & \makecell{Adapter\\training} & \makecell{Adapter\\steps} & \makecell{Adapter\\parameters} \\
\midrule
$500$ & $0.0014$\,s & $16.6$\,s & $600$ & $1.25$\,M \\
$1{,}000$ & $0.0018$\,s & $16.2$\,s & $620$ & $1.25$\,M \\
$2{,}000$ & $0.0033$\,s & $16.2$\,s & $620$ & $1.25$\,M \\
$4{,}000$ & $0.0063$\,s & $19.4$\,s & $750$ & $1.25$\,M \\
$8{,}000$ & $0.0123$\,s & $41.5$\,s & $1{,}500$ & $1.25$\,M \\
$32{,}463$ & $0.0484$\,s & $160.5$\,s & $6{,}084$ & $1.25$\,M \\
\bottomrule
\end{tabular}
\end{table}

\FloatBarrier

\section{A mask-supervised reference}
\label{sec:s-sam3}

SAM~3~\cite{sam3} segments every instance of a named concept. Unlike CAPTOR, which
counts captions over a frozen encoder, it is trained with gradients on millions of
labelled masks, carries its own text encoder and uses far more data. It is
therefore a reference rather than a baseline. Table~\ref{tab:sam3} runs the
released model on a fixed subset of every segmentation benchmark, with CAPTOR
scored on the same images. SAM~3 leads on the four natural-image benchmarks. In
pathology it finds nuclei but no tissue regions and names every tile
``lymphocyte'', so CAPTOR leads on breast and lung. SAM~3's training data include
fifteen external datasets with instance masks that its paper does not name, so
overlap with the COCO, VOC and Cityscapes images cannot be ruled out. This
suggests that in specialized domains, where direct text-image labels are often
unavailable, CAPTOR shows its advantage.

\begin{table}[!htp]\centering
\caption{How the released SAM~3, a model trained on millions of labelled masks, compares
with CAPTOR on the same images. Each benchmark uses a fixed part of its reporting half. That is $250$ images drawn
with seed $0$ on COCO, VOC and COCO-Object, all $250$ reporting images on Cityscapes,
and every reporting tile in pathology. Each keeps the
protocol and CAPTOR setting of its table in the paper, and SAM~3 uses its own released
decision rule. \emph{Scored} says what each protocol scores: \emph{objects} are grid
cells covered by one object, \emph{all pixels} include background, and
\emph{tissues} are grid cells of annotated tissue. The two are not comparable models,
as Section~\ref{sec:s-sam3} explains. The last column is SAM~3 minus CAPTOR, with
$95\%$ intervals over images, or patients in pathology. SAM~3 leads on every
natural-image benchmark and fails in pathology, where it names every tile
``lymphocyte''.}
\label{tab:sam3}
\begin{fitwidth}
\setlength{\tabcolsep}{4pt}
\begin{tabular}{llcccc}
\toprule
Benchmark & Scored & Images & CAPTOR & SAM~3 & \makecell{SAM~3\\ $-$ CAPTOR} \\
\midrule
COCO & Objects & $250$ & $0.517$ & $\mathbf{0.753}$ & \makecell{$+0.236$\\ $[+0.190, +0.288]$} \\
Pascal VOC & Objects & $250$ & $0.820$ & $\mathbf{0.951}$ & \makecell{$+0.131$\\ $[+0.093, +0.176]$} \\
COCO-Object & All pixels & $250$ & $0.304$ & $\mathbf{0.609}$ & \makecell{$+0.306$\\ $[+0.265, +0.338]$} \\
Cityscapes & All pixels & $250$ & $0.559$ & $\mathbf{0.704}$ & \makecell{$+0.145$\\ $[+0.115, +0.212]$} \\
\midrule
BCSS (breast) & Tissues & \makecell{$134$ tiles\\ $62$ patients} & $\mathbf{0.521}$ & $0.023$ & \makecell{$-0.498$\\ $[-0.556, -0.436]$} \\
\makecell[l]{LUAD-HistoSeg\\ (lung)} & Tissues & \makecell{$367$ tiles\\ $15$ patients} & $\mathbf{0.325}$ & $0.020$ & \makecell{$-0.305$\\ $[-0.361, -0.239]$} \\
\bottomrule
\end{tabular}
\end{fitwidth}
\end{table}

\FloatBarrier

\FloatBarrier


\end{document}